%% file: paper-icml23.tex
\documentclass[nohyperref,accepted,usenames,dvipsnames]{article}
\usepackage[dvipsnames]{xcolor}
\usepackage{microtype}
\usepackage{graphicx}
\usepackage{subfigure}
\usepackage{booktabs}
\usepackage{arydshln}
\usepackage{siunitx}
\usepackage{hyperref}

\usepackage{icml2023}

\usepackage{amsmath}
\usepackage{amssymb}
\usepackage{mathtools}
\usepackage{amsthm}

\usepackage[capitalize,noabbrev]{cleveref}
\usepackage{researchpack}
\usepackage{enumitem}
\usepackage{xspace}
\usepackage{epsdice}
\usepackage{multirow}
\usepackage{systeme}

\hypersetup{
    colorlinks=true,
    citecolor=blue,
    linkcolor=blue,
    filecolor=blue,
    urlcolor=blue,
}

\definecolor{ClevrGray}{RGB}{128,128,128}
\definecolor{ClevrGreen}{RGB}{35,127,21}
\definecolor{ViolinBlue}{RGB}{34,113,144}
\definecolor{ViolinRed}{RGB}{226,94,162}

\newcommand{\changed}[1]{\textcolor{black}{#1}}

\newcommand{\acronym}{COncept-level cOntinual Learning\xspace}
\newcommand{\method}{\textsc{cool}\xspace}

\newcommand{\naive}{\textsc{na\"ive}\xspace}
\newcommand{\restart}{\textsc{restart}\xspace}

\newcommand{\er}{\textsc{er}\xspace}

\newcommand{\der}{\textsc{der}\xspace}
\newcommand{\derpp}{\textsc{der++}\xspace}

\newcommand{\lwf}{\textsc{lwf}\xspace}
\newcommand{\ewc}{\textsc{ewc}\xspace}

\newcommand{\offline}{\textsc{offline}\xspace}

\newcommand{\CBM}{\textsc{CBM}\xspace}
\newcommand{\CBMs}{\textsc{CBM}s\xspace}
\newcommand{\DeepProbLog}{DeepProbLog\xspace}

\newcommand{\MNISTAdd}{{\tt MNIST-Addition}\xspace}
\newcommand{\MNISTSeq}{{\tt MNAdd-Seq}\xspace}
\newcommand{\MNISTShortcut}{{\tt MNAdd-Shortcut}\xspace}
\newcommand{\CLEFVR}{{\tt CLE4EVR}\xspace}

\newcommand{\MZero}{\includegraphics[width=1.85ex]{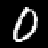}\xspace}
\newcommand{\MOne}{\includegraphics[width=1.85ex]{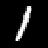}\xspace}
\newcommand{\MTwo}{\includegraphics[width=1.85ex]{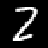}\xspace}
\newcommand{\MThree}{\includegraphics[width=1.85ex]{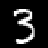}\xspace}
\newcommand{\MFour}{\includegraphics[width=1.85ex]{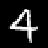}\xspace}
\newcommand{\MFive}{\includegraphics[width=1.85ex]{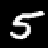}\xspace}
\newcommand{\MSix}{\includegraphics[width=1.85ex]{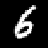}\xspace}
\newcommand{\MSeven}{\includegraphics[width=1.85ex]{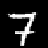}\xspace}
\newcommand{\MEight}{\includegraphics[width=1.85ex]{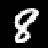}\xspace}
\newcommand{\MNine}{\includegraphics[width=1.85ex]{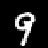}\xspace}

\newcommand{\task}{\ensuremath{\calT}\xspace}
\newcommand{\dataset}{\ensuremath{\calD}\xspace}
\newcommand{\memory}{\ensuremath{\calM}\xspace}
\newcommand{\BK}{\ensuremath{\mathsf{K}}\xspace}

\newcommand{\KL}{\ensuremath{\mathsf{KL}}\xspace}

\newtheorem{example}{Example}
\newtheorem{theorem}{Theorem}
\newtheorem{lemma}{Lemma}

\icmltitlerunning{Neuro-Symbolic Continual Learning}

\begin{document}

\twocolumn[
\icmltitle{Neuro-Symbolic Continual Learning:\\ Knowledge, Reasoning Shortcuts and Concept Rehearsal}



\icmlsetsymbol{equal}{*}

\begin{icmlauthorlist}
\icmlauthor{Emanuele Marconato}{equal,unipi,disi}
\icmlauthor{Gianpaolo Bontempo}{equal,unipi,unimore}
\icmlauthor{Elisa Ficarra}{unimore}
\icmlauthor{Simone Calderara}{unimore}
\icmlauthor{Andrea Passerini}{disi}
\icmlauthor{Stefano Teso}{cimec,disi}
\end{icmlauthorlist}

\icmlaffiliation{disi}{DISI, University of Trento, Italy}
\icmlaffiliation{unimore}{University of Modena and Reggio Emilia, Italy}
\icmlaffiliation{cimec}{CIMeC, University of Trento, Italy}
\icmlaffiliation{unipi}{University of Pisa, Italy}

\icmlcorrespondingauthor{Emanuele Marconato}{emanuele.marconato@unitn.it}

\icmlkeywords{Continual Learning, Neuro-symbolic Integration, Concepts, Rehearsal}

\vskip 0.3in
]



\printAffiliationsAndNotice{\icmlEqualContribution} 

\begin{abstract}
    We introduce Neuro-Symbolic Continual Learning,
    where a model has to solve \textit{a sequence of neuro-symbolic tasks}, that is, it has to map sub-symbolic inputs to high-level concepts and compute predictions by \textit{reasoning} consistently with prior knowledge.
    Our key observation is that neuro-symbolic tasks, although different, often share concepts whose \textit{semantics} remains stable over time.
    Traditional approaches fall short:  existing continual strategies ignore knowledge altogether, while stock neuro-symbolic architectures suffer from catastrophic forgetting.
    We show that leveraging prior knowledge by combining neuro-symbolic architectures with continual strategies \textit{does} help avoid catastrophic forgetting, but also that doing so can yield models affected by \textit{reasoning shortcuts}.  These undermine the semantics of the acquired concepts, even when detailed prior knowledge is provided upfront and inference is exact, and in turn continual performance.
    To overcome these issues, we introduce \method, a \textbf{CO}ncept-level c\textbf{O}ntinual \textbf{L}earning strategy tailored for neuro-symbolic continual problems that acquires high-quality concepts and remembers them over time.
    Our experiments on three novel benchmarks highlights how \method attains sustained high performance on neuro-symbolic continual learning tasks in which other strategies fail.\footnote{The data and code are available at \hyperlink{https://github.com/ema-marconato/NeSy-CL}{https://github.com/ema-marconato/NeSy-CL} }
\end{abstract}

\section{Introduction}
\label{sec:introduction}

We initiate the study of Neuro-Symbolic Continual Learning (NeSy-CL), in which the goal is to \textit{solve a sequence of neuro-symbolic tasks}.
As is common in neuro-symbolic (NeSy) prediction~\citep{manhaeve2018deepproblog,xu2018semantic,giunchiglia2020coherent,hoernle2022multiplexnet,ahmed2022semantic}, the machine is provided \textit{prior knowledge} relating one or more target labels to symbolic, high-level concepts \textit{extracted} from sub-symbolic data, and has to compute a prediction by \textit{reasoning} over said concepts.
The central challenge of Nesy-CL is that the data distribution and the knowledge may vary across tasks.
E.g., in medical diagnosis knowledge may encode known relationships between possible symptoms and conditions, while different tasks are characterized by different distributions of X-ray scans, symptoms and conditions.
The goal, as in continual learning (CL)~\citep{parisi2019continual}, is to obtain a model that \textit{attains high accuracy on new tasks without forgetting what it has already learned} under a limited storage budget.

\begin{figure*}[!t]
    \centering
    \includegraphics[height=10em]{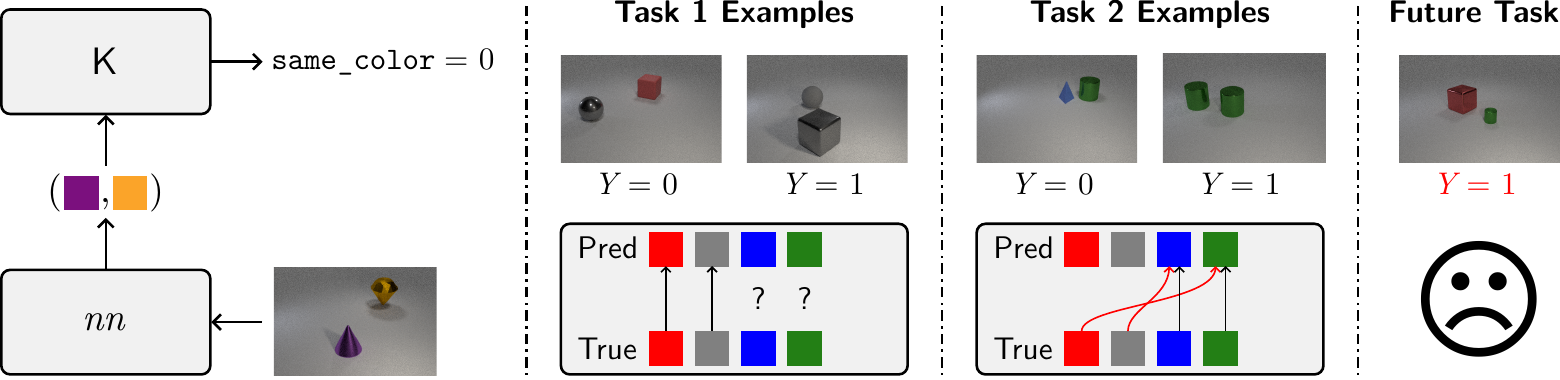}
    \vspace{-1em}
    \caption{
    \textbf{Left}: DeepProbLog extracts concepts $\vc$ from a sub-symbolic input $\vx$ and reasons over \changed{prior knowledge \BK provided upfront} to obtain a prediction $\vy$.
    \textbf{Right}:  Simplified illustration of our \CLEFVR benchmark, \changed{restricted to} color concepts only. \changed{The full setting is reported in \cref{sec:experiments}.} The goal is to predict whether two objects have the $Y = {\tt same\_color}$.
    The first task includes examples of \textbf{\textcolor{ClevrGray}{gray}} and \textbf{\textcolor{red}{red}} objects only, and the model classifies them by learning the intended mapping between input colors and concepts.
    The second one includes \textbf{\textcolor{ClevrGreen}{green}} and \textbf{\textcolor{blue}{blue}} objects only, and the model can learn a \textit{reasoning shortcut} mapping the four colors to two concepts, achieving high accuracy on both tasks but compromising performance on \textit{future} tasks.
    This is exactly the issue addressed by our approach, \method.
    }
    \label{fig:main}
    \vspace{-1.5em}

\end{figure*}

Existing approaches are insufficient for NeSy-CL:
neuro-symbolic models are designed for offline learning and as such suffer from \textit{catastrophic forgetting}~\citep{parisi2019continual}, while continual learning strategies \changed{are designed for neural networks that neglect \changed{prior knowledge}, preventing applications to tasks where compliance with regulations is key, e.g., safety critical tasks~\citep{ahmed2022semantic,hoernle2022multiplexnet}}.
It is tempting to tackle NeSy-CL by pairing a SotA neuro-symbolic architecture, such as DeepProbLog~\citep{manhaeve2018deepproblog}, with a proven rehearsal or distillation strategy, for instance dark experience replay~\citep{buzzega2020dark}.  This yields immediate benefits, in the sense that prior knowledge makes the model more robust to catastrophic forgetting, as we will show.
However, we show also that it is flawed, because it cannot prevent the model from acquiring \textit{reasoning shortcuts} (defined in \cref{sec:shortcuts}), through which it attains high task accuracy by acquiring unintended concepts with \textit{task-specific semantics}, as illustrated in \cref{fig:main}.
In turn, reasoning shortcuts entail poor cross-task transfer.

Our key observation is that, even though neuro-symbolic tasks may differ in terms of knowledge and distribution, \textit{the semantics of the concepts they rely on must remain stable over time}.
For instance, in automated protein annotation inferred signal peptides entail a catalytic function~\citep{rost2003automatic} regardless of the specific protein under examination, and in vehicle routing presence of pedestrians on the road is sufficient to rule out certain routes~\citep{xu2020explainable} regardless of location and weather conditions.

Prompted by this insight, we propose \method, a simple but effective \acronym strategy, that aims at \textit{acquiring high-quality concepts and preserving them across tasks}.
\method makes use of a small amount of concept supervision to acquire high-quality concepts and explicitly preserves them with a \textit{concept} rehearsal strategy, avoiding reasoning shortcuts in all tasks.
\method is applicable to a variety of NeSy architectures, and -- as shown by our experiments with DeepProbLog~\citep{manhaeve2018deepproblog} and Concept-based Models~\citep{koh2020concept} -- easily outperforms state-of-the-art continual strategies on three novel, challenging NeSy-CL problems, achieving better concept quality and predictive accuracy on past and OOD tasks.

\noindent
\textbf{Contributions.}  Summarizing, we:
\begin{enumerate}[leftmargin=1.25em]

    \item Introduce neuro-symbolic continual learning as a novel and challenging machine learning problem.\
    
    \item We show that knowledge readily improves forgetting in some scenarios, but also that it is insufficient to prevent reasoning shortcuts -- which worsen forgetting and compromise transfer to new tasks -- in others.
    \item Propose new NeSy-CL benchmarks for evaluating continual performance with and without reasoning shortcuts.
    \item Introduce \method, a novel continual strategy that supports identifying concepts with the intended semantics and preserves them across tasks. 
    
    \item Show empirically that \method outperforms state-of-the-art continual strategies on these challenging benchmarks.

\end{enumerate}

\section{Neuro-Symbolic Continual Learning}
\label{sec:preliminaries}

\noindent
\textbf{Notation.}  Throughout, we indicate scalar constants $x$ in lower-case, random variables $X$ in upper case, and ordered sets of constants $\vx$ and random variables $\vX$ in bold typeface.
The symbol $[n]$ stands for the set $\{1, \ldots, n\}$ and $\vx \models \BK$ indicates that $\vx$ satisfies a logical formula $\BK$.
We say that a distribution $p(\vA \mid \vB ; \BK)$ is \textit{consistent} with $\BK$, written $p \models \BK$, if it holds that $p(\va \mid \vb ; \BK) > 0$ implies $(\va, \vb) \models \BK$, \ie if $p$ associates zero mass to all states that violate \BK.

\subsection{Problem Statement}

We are concerned with solving a \textit{sequence of neuro-symbolic prediction tasks}, each requiring to learn a classifier mapping a (partially) sub-symbolic input $\vx \in \bbR^d$ to $n \ge 1$ labels $\vy \in \bbN^n$.
What makes them neuro-symbolic is that:
%
%
(\textit{i}) The labels $\vy$ depend entirely on the state of $k$ \textit{symbolic concepts} $\vc = (c_1, \ldots, c_k)^\top$ capturing high-level aspects of the input $\vx$.
(\textit{ii}) The concepts $\vc$ depend on the \textit{sub-symbolic input} $\vx$ in an intricate manner and are best extracted using deep learning techniques.
(\textit{iii}) The way in which the labels $\vy$ depend on the concepts $\vc$ is specified by \textit{\changed{prior knowledge}} $\BK$, necessitating reasoning during inference.
%

We make the natural assumption that the \textit{semantics of the concepts appearing in the various tasks remain constant over time}.  This assumption lies at the heart of knowledge representation and ontology design, where concepts serve as a \textit{lingua franca} for the exchange and reuse of knowledge across application boundaries~\citep{gruber1995toward} and with human stakeholders~\citep{kambhampati2021symbols}.

Formally, each task $t \in \bbN$ is defined by a data generating distribution $p^{(t)}(\vX, \vC, \vY ; \BK^{(t)})$ that factorizes as:
\[
    p^{(t)}(\vY \mid \vC; \BK^{(t)}) \cdot p(\vC \mid \vX) \cdot p^{(t)}(\vX)
    \label{eq:generative-process}
\]
Here, $\BK^{(t)}$ denotes the knowledge relevant to the $t$-th task.\footnote{The knowledge might depend also on discrete variables in $\vX$;  we suppress this dependency in the notation for readability.}
As customary in NeSy, we assume the knowledge to correctly describe the ground-truth generative process and that, therefore, the label distributions are \textit{consistent} with their respective prior knowledge, \ie $p^{(t)}(\vY \mid \vX; \BK^{(t)}) \models \BK^{(t)}$ for all $t$.
The key feature of \cref{eq:generative-process} is that $p(\vC \mid \vX)$ does not depend on $t$, capturing our assumption that concept semantics are stable.  For instance, if $C_{\tt dog}$ represents the notion of ``dog'', then $p(C_{\tt dog} \mid \vX)$ only depends on whether an image $\vx$ in fact depicts a dog, regardless of context, style, likelihood of observing a dog, and role of dogs in determining the label.  See
\cref{sec:semantics} for an in-depth discussion.
Critically, the distribution of observed inputs, concepts and labels, and the knowledge \textit{are allowed to differ between tasks}.
This means that, \eg known concepts may stop occurring, play different roles in $\BK^{(t+1)}$ than they did in $\BK^{(t)}$, and entirely new concepts may appear.

\textbf{Handling catastrophic forgetting.}  At step $t$, the machine obtains a task $\task^{(t)} = (\dataset^{(t)}, \BK^{(t)})$ consisting of a data set $\dataset^{(t)}$, sampled i.i.d. according to \cref{eq:generative-process}, and knowledge $\BK^{(t)}$.
The goal is to find parameters $\theta$ that achieve low \textit{average risk} over all tasks observed so far, defined as:
\begin{align}
    \calL(\theta, \task^{1:t})
        & = \frac{1}{t} \sum_{s \in [t]} \calL(\theta, \task^{(s)})
    \\
        & = \frac{1}{t} \calL(\theta, \task^{(t)}) + \frac{t - 1}{t} \calL(\theta, \task^{1:(t-1)})
        \label{eq:average-risk}
\end{align}
where $\task^{1:t} = \{ \task^{(1)}, \ldots, \task^{(t)} \}$ is the collection of all tasks observed so far, $\calL(\theta, \task) := \frac{1}{|\dataset|} \sum_{(\vx, \vy) \in \dataset} \ell(\theta, (\vx, \vy); \BK)$, and $\ell$ is a loss function over $\vy$.

As in regular continual learning, we assume storage is insufficient to hold data from all tasks, meaning that $\calL(\theta, \task^{1:(t-1)})$ cannot be evaluated exactly.
Ignoring this term, as done by offline approaches, leads to \textit{catastrophic forgetting}:  by focusing on the loss of the current task, models tend to forget the information necessary to solve the previous tasks.
%
CL algorithms mitigate forgetting using strategies like rehearsal~\citep{buzzega2020dark,parisi2019continual,delange2021continual,boschini2022class,rebuffi2017icarl} (\ie replaying few examples of previous tasks), regularization~\citep{huszar2018note,li2017learning} (\ie slowing down parameter shift through additional terms in the loss function), or architectural modifications~\citep{rusu2016progressive} (\ie freezing and adding new parameters at each task).  See \cref{sec:related-work} for an overview.

Importantly, all these strategies focus on \textit{optimizing the accuracy on the labels only}.  This is sensible in CL but, as shown in \cref{sec:shortcuts}, insufficient in NeSy-CL.

\subsection{DeepProbLog}
\label{sec:deepproblog}

DeepProbLog~\citep{manhaeve2018deepproblog} is a state-of-the-art model ideally suited to solve tasks of the form in \cref{eq:generative-process}.
It decomposes prediction into two steps, cf. \cref{fig:main} (left).
At the lower level, it implements each concept as a Boolean or categorical random variable $C_j$, whose distribution $p_\theta(C_j \mid \vx)$ is flexibly parameterized by a neural network $nn_j(\vx; \theta)$.
This implies that concepts are mutually independent given the input, \ie $p_\theta(\vC \mid \vx) = \prod_{j \in [k]} p_\theta(C_j \mid \vx)$.
%
At the upper level, it models the distribution of labels conditioned on concepts as a \textit{uniform distribution} over the support of \BK, and specifically as:\footnote{Non-uniform distributions consistent with the knowledge can also be modelled, as done for instance by \citet{ahmed2022semantic}.}
\[
    u_{\BK}(\vy \mid \vc) = \frac{1}{Z(\vc; \BK)} \cdot \Ind{(\vc, \vy) \models \BK}
    \label{eq:deepproblog-rules}
\]
where $Z(\vc; \BK) = \sum_{\vy} \Ind{ (\vc, \vy) \models \BK }$ is a normalization constant.
The overall label distribution is obtained by marginalizing over $\vC$:
\[
    \textstyle
    p_\theta(\vy \mid \vx; \BK)
        = \sum_{\vc} u_{\BK}(\vy \mid \vc) \cdot \prod_{j \in [k]} p_\theta(c_j \mid \vx)
    \label{eq:deepproblog-joint}
\]
Since the indicator function in \cref{eq:deepproblog-rules} evaluates to zero for all values of $\vc$ that violate \BK, the label distribution in \cref{eq:deepproblog-joint} is \textit{by construction} consistent with $\BK$.


\begin{example}
\textit{\MNISTAdd \citep{manhaeve2018deepproblog} is a prototypical neuro-symbolic task that requires learning a mapping from pairs of MNIST digits $\vx = (\vx_1, \vx_2)$ to their sum $Y$, \eg from $\vx = (\MFour, \MOne)$ to $y = 5$.
It can be readily modelled in DeepProbLog using two concepts $C_1$ and $C_2$ ranging in $\{ 0, \ldots, 9 \}$, each predicted by a convolutional network $nn(\vx_j)$, and a constraint $\BK = (C_1 + C_2 =  Y)$.}
\end{example}

Given a task $\task = (\calD, \BK)$, the parameters $\theta$ are usually learned by maximizing the log-likelihood $\calL(\theta, \dataset) := \frac{1}{|\dataset|} \sum_{(\vx, \vy) \in \dataset} \log p_\theta(\vy \mid \vx; \BK)$ via stochastic gradient descent.
Computing the (gradient of the) likelihood $p_\theta(\vy \mid \vx)$ and the most likely prediction $\widehat{\vy} \in \argmax_{\vy} p_\theta(\vy \mid \vx; \BK)$ requires to evaluate $Z$, which is intractable in general.
To make inference practical, DeepProbLog exploits knowledge compilation~\citep{darwiche2002knowledge} to convert the distribution $u_{\BK}$ into a probabilistic circuit.  Once in this format, the above operations take time linear in the size of the circuit~\citep{choi2020probabilistic,vergari2021compositional}.

In the remainder of the paper, we focus on DeepProbLog as it offers a sound probabilistic architecture and exact inference.  Our results, however, do transfer to many other neuro-symbolic architectures, as discussed in~\cref{sec:related-work}.

\section{Knowledge and Reasoning Shortcuts}
\label{sec:shortcuts}

All neuro-symbolic architectures, including DeepProbLog, are designed for \textit{offline} settings, and as such they easily fall prey of catastrophic forgetting when applied to NeSy-CL problems.
This issue is illustrated by the following example and demonstrated empirically in \cref{sec:experiments}.

\begin{example}
\label{example:mnist-seq}
\textit{We introduce \MNISTSeq, a continual extension of \MNISTAdd in which tasks differ in what digits are observed.
Specifically, each task $t = 0, \ldots, 8$ consists of all pairs of digits $\vx = (\vx_1, \vx_2)$ whose sum is either $2t$ or $2t + 1$.  By construction, sums above $9$ cannot be obtained by adding smaller digits, so these no longer occur in later tasks.
Since DeepProbLog maximizes the likelihood of the current task, it quickly forgets small digits at $t \ge 5$ and can no longer classify sums involving them correctly.}
\end{example}

\subsection{Knowledge Helps Remembering \ldots}
\label{sec:knowledge-helps}

A natural first step toward solving NeSy-CL is to \changed{bundle a NeSy predictor -- say, DeepProbLog -- with any state-of-the-art CL strategy.  Focusing on experience replay, doing so amounts to storing a handful of well chosen labeled (or predicted) examples $(\vx, \vy)$ from past tasks $1, \ldots, t-1$, and replaying them when fitting DeepProbLog on the current task $t$ together with the corresponding prior knowledge $\BK^t$.}\footnote{The only non-trivial aspect is that, in addition to the replay buffer, we also has to store the past knowledge, so as to ensure be able to match the updated concepts with the past labels.}

Doing so immediately brings a number of benefits.
First and foremost, the knowledge encodes the valid, stable relationship between the concepts and the labels to be prediction.
This implies that predicted concepts can be always correctly mapped to a corresponding label, and that this inference step is immune from forgetting.  This is especially significant considering that the top layers of neural networks are those most affected by catastrophic forgetting~\citep{wu2019large}.
This effect is clearly visible in our experiments, cf. \cref{sec:experiments}.

Conversely, prior knowledge effectively reduces the space of candidate concepts, providing further guidance to the model.  If it reduces the space to only those having the intended semantics, then this simple setup can be very effective at tackling NeSy-CL problems.

\subsection{\ldots But Does Not Prevent Reasoning Shortcuts}

In general, however, this setup is insufficient.
The core issue is that knowledge might not be enough to identify the right
concept distribution $p(\vC \mid \vX)$ using label annotations alone, in the sense that -- depending on how the knowledge and training data are structured -- it may be possible \changed{(in both offline and continual settings)} to \textit{correctly classify all training examples even using concepts with unintended semantics}.  We refer to these situations as \textit{reasoning shortcuts}.

This intuition is formalized in \cref{thm:ll-optima-satisfy-bk}.
Here, we write $\Theta$ to indicate the set of all possible parameters of (the neural networks implementing) $p_\theta(\vC \mid \vX)$, and $\Theta^*(\BK, \dataset) \subseteq \Theta$ for the parameters that maximize the log-likelihood $\calL(\theta, \dataset)$.  Also, $\BK[\vV/\vv]$ is the knowledge obtained by substituting all occurrences of variables $\vV$ with constants $\vv$.  For instance, in \MNISTAdd $\BK[Y/2]$ amounts to $C_1 + C_2 = 2$.

\begin{theorem}
\label{thm:ll-optima-satisfy-bk}
A model with parameters $\theta$ attains maximal likelihood, \ie $\theta \in \Theta^*({\BK}, \dataset)$, if and only if, for all $(\vx, \vy) \in \dataset$, it holds that $p_\theta(\vC \mid \vx) \models \BK[\vY/\vy]$.
\end{theorem}

All proofs can be found in \cref{sec:proofs}.
\cref{thm:ll-optima-satisfy-bk} states that, as long as the concept distribution output by the learned neural network satisfies the knowledge for each training example, the log-likelihood is maximal.\footnote{This theorem essentially shows that, from the neural network's perspective, the reasoning layer of DeepProbLog has the same effect as the Semantic Loss~\citep{xu2018semantic}.}
The ground-truth concept distribution $p(\vC \mid \vX)$ is a possible solution, but it is not necessarily \textit{the only one}.
In this case, fitting DeepProbLog -- and indeed any NeSy approach that optimizes for label accuracy only -- \textit{does not guarantee that the learned concepts have the correct semantics}.
To see this, consider the following example.


\begin{example}
\label{example:shortcut}
\textit{Consider \MNISTAdd and take a subset $\dataset$ including only pairs of examples of four possible sums: $\MZero + \MSix = 6$, $\MFour + \MSix = 10$, $\MTwo + \MEight = 10$, and $\MFour + \MEight = 12$.  Then, there exist many concept distributions that satisfy the knowledge on all examples, including:
\begin{align*}
    \MZero \mapsto 0, \quad\MTwo \mapsto 2, \quad\MFour \mapsto 4, \quad\MSix \mapsto 6, \quad\MEight \mapsto 8
    \\
    \MZero \mapsto 5, \quad\MTwo \mapsto 7, \quad\MFour \mapsto 9, \quad\MSix \mapsto 1, \quad\MEight \mapsto 3
\end{align*}
where the remaining concepts are allocated arbitrarily and $\vx \mapsto c$ is a shorthand for $p(C = c \mid \vx) = \Ind{C = c}$.
Only the first distribution has the intended semantics, whereas the second one is a reasoning shortcut.  We remark that DeepProbLog does acquire this shortcut in practice, as illustrated by our experiments.  \cref{sec:shortcut-solutions} explains how shortcuts emerge in the data sets used in our experiments.}
\end{example}

\changed{Notice that the Theorem applies to both offline learning (\ie $\dataset$ is fixed) and NeSy-CL (\ie $\dataset$ indicates the training set of any given task).  Yet, reasoning shortcuts are especially impactful in the latter.}
This is exemplified in \cref{fig:main} (right).  Here, DeepProbLog has learned high-quality concepts to solve the first task, but quickly forgets them when solving the second task, precisely because it falls pray of a reasoning shortcut that achieves high training and rehearsal accuracy on both tasks by satisfying the knowledge using concepts with unintended semantics.
We provide additional concrete examples in \cref{sec:shortcut-solutions}.
In turn, reasoning shortcuts can dramatically affect forgetting and performance on future and OOD NeSy tasks, as shown by our experiments.




\section{Addressing NeSy-CL with \method}
\label{sec:method}

To this end, we introduce \method, a \acronym that acquires concepts with the intended semantics and preserves them over time, attaining sustained high performance.
Formally, \method is designed to satisfy two desiderata: (D1) $p_\theta(\vC \mid \vX)$ should quickly approximate $p(\vC \mid \vX)$, and (D2) $p_\theta(\vC \mid \vX)$ should remain stable across tasks.
D2 is straightforward, however we stress that it is only meaningful if D1 also holds:  unless the learned concepts are high-quality, there is little benefit in remembering them.

In order to comply with D1, \method makes use of a small number of densely annotated examples to quickly identify high-quality concepts, which -- as we have shown -- cannot always be guaranteed using knowledge alone.  In practice, an average cross-entropy is added to the loss of these examples.
To cope with D2, \method implements a novel \textit{concept rehearsal} strategy that stabilizes $p_\theta(\vC \mid \vX)$ across tasks.  This is motivated by the fact that concept stability helps to upper bound the average risk.
Specifically,

\begin{theorem}
\label{thm:concept-distillation-is-enough}
Consider tasks $\task^{1:t}$.  If the current model $\theta$ and the past one $\theta^{(t-1)}$ assign non-zero likelihood to all examples in $\dataset^{1:t}$, there exists a finite constant $\gamma$, depending only on the model architecture, knowledge and data, such that the average risk in \cref{eq:average-risk} is upper bounded by:
\begin{align}
    & \frac{1}{t} \calL(\theta, \dataset^{(t)}) + \frac{t - 1}{t}
        \Big[
            \calL(\theta^{(t-1)}, \dataset^{1:(t-1)})
    \label{eq:avg-risk-ub}
    \\
    & \quad + \gamma \sum_{s \le t} \sum_{(\vx, \vy) \in \dataset^{(s)}} \norm{
                p_{\theta}(\vC \mid \vx) - p_{\theta^{(t-1)}}(\vC \mid \vx)
            }_1
        \Big]
    \nonumber
\end{align}
\end{theorem}

\vspace{-0.6em}

In words, this means that if the past model $\theta^{(t-1)}$ performs well on all past tasks (\ie the middle term in \cref{eq:avg-risk-ub} is small), a new model that performs well on the current task (the first term is small) and whose concept distribution is close to that of the old model (the last term is small), also performs well on past tasks (the average risk in \cref{eq:average-risk} is small).
Critically, this results holds regardless of how the prior knowledge $\BK^{1:t}$ of the various task is chosen.

\method implement this requirement by combining the original training loss with an extra penalty $\calL_{\method}$, defined as: 
\begin{align}
    \calL_\mathrm{\method} :=
          \frac{1}{|\memory|} \sum_{(\vx, \Tilde{\vq}_c, \vy) \in \memory} \big[ &\alpha \cdot \KL \big( p_\theta(\vC \mid \vx) \, \| \, \Tilde{\vq}_c) 
          \label{eq:cool}
          \\
        & - \beta \cdot \log p_\theta(\vY = \vy \mid \vx; \BK^{(t)}) \big]
        \nonumber
\end{align}
Here, $\memory$ denotes \changed{the mini-batch of examples extracted from the replay buffer}, $\alpha$ denotes the scalar weight associated to the concept-reharsal strategy, and $\beta$ the weight of the replay strategy on $\vy$. The $\KL$ term is evaluated between the predicted concept distribution and the stored one $\Tilde{\vq}_c = p_{\theta^{(t-1)}}(\vC \mid \vx)$.
Notice that, by Pinsker's inequality, the KL upper bounds the (square of the) $L_1$ distance, meaning that \method indirectly optimizes the bound in \cref{eq:avg-risk-ub}.

\subsection{Benefits and Limitations}

\method is explicitly designed to acquire high-quality concepts and retain them across tasks by combining knowledge, concept rehearsal, and a modicum of concept supervision.  This substantially improves performance on past, future, and OOD tasks sharing these concepts, as demonstrated in \cref{sec:experiments}.
\method works even if the knowledge $\BK^{(t)}$ changes across tasks
%
and new concepts appear over time:  these can be encoded as additional neural predicates in DeepProbLog, and \method will take care of remembering the known concepts while leaving room to learn the new ones.

One limitation of \method is that, in the general case, it requires a handful of densely annotated examples.
The same requirement can be found in other settings where concept quality is critical.
For instance, concept supervision is key in concept-based models -- which strive to generate concept-level explanations for their predictions -- to ensure the acquired concepts are \textit{interpretable}~\citep{koh2020concept,chen2020concept,marconato2022glancenets}.
It is also a prerequisite for guaranteeing that learned representations acquired by general (deep) latent variable models are \textit{disentangled}, as shown theoretically~\citep{locatello2019challenging} and empirically~\citep{locatello2020disentangling}.
We stress that concept supervision is \textit{not} required if knowledge and data disallow shortcut solutions, as is the case in \MNISTSeq (see \cref{sec:experiments}), although it does help avoiding sub-optimal parameters even in this case.
If reasoning shortcuts \textit{are} possible, however, concept supervision becomes essential, because -- by construction -- knowledge and labels alone are insufficient to pin down the correct semantics, hindering concept quality.
Moreover, in many situations, annotating just \textit{some} concepts is sufficient to rule out reasoning shortcuts.

\section{Empirical Evaluation}
\label{sec:experiments}

We address empirically the following research questions:
\begin{itemize}[leftmargin=2em]

    \item[\textbf{Q1}:]  Does knowledge help to stabilize the continual learning process and reduce the need for supervision?

    \item[\textbf{Q2}:]  Does \method help avoid reasoning shortcuts when knowledge alone fails, thus facilitating past and future continual performance?

    \item[\textbf{Q3}:]  How much concept supervision does \method need?

\end{itemize}
To answer these questions, we compared \method against several representative continual strategies on three \textit{novel} and challenging NeSy-CL benchmarks.  Additional results and details on data sets, metrics, and hyperparameters can be found in the Appendices.

\textbf{Data sets.}  Existing NeSy and CL benchmarks are designed for offline settings or lack any sort of \changed{prior knowledge}, respectively.  Hence, in order to evaluate \method, we introduce three novel NeSy-CL benchmarks specifically designed to evaluate impact of knowledge, concept quality and robustness to reasoning shortcuts, briefly described next.

\underline{\MNISTSeq} is the problem introduced in \cref{example:mnist-seq} and it is designed \textit{not} to contain reasoning shortcuts.
In short, inputs $\vx$ are pairs of MNIST~\citep{lecun1998mnist} digits labeled with their sum $y$; each digit is mapped to a concept, and \BK specifies that their sum must match the label.
In each task $t = 0, \ldots, 8$ includes only examples with labels $y \in \{ 2 \cdot t, 2 \cdot t + 1 \}$, making this problem both \textit{label incremental} (only two out of $18$ possible labels are observed per task) and \textit{concept incremental} (higher digits only appear in later tasks, while lower digits disappear, see \cref{sec:app-datasets}).  The data set holds $42$k training examples, of which we used $8.4$k for validation, and $6$k test examples.

\underline{\MNISTShortcut} is a simple two-task version of \MNISTAdd used here to illustrate the impact of \textit{reasoning shortcuts}.
The first task includes only even digits and the second one only odd ones.  In the first task we include 4 types of examples: (i) $\MZero + \MSix = 6$, (ii) $\MFour + \MSix = 10$, (iii) $\MTwo + \MEight = 10$, and (iv) $\MFour + \MEight = 12$.  In the second task, we allow all possible sums of odd digits $\{ \MOne, \MThree, \MFive, \MSeven, \MNine \}$.  As shown in Example~\ref{example:shortcut} and further discussed in \cref{sec:shortcut-solutions}, the four sums in the first task are not sufficient to identify the correct digits, \ie different shortcuts are possible.
This data consists of $13.8$k examples, $2.8$k of which are reserved for validation  and $2$k for testing.
We also include an additional OOD test set with $4$k unseen combinations of all concepts, like sums involving an odd and an even digit, allowing us to probe the efficacy on a plausible future task.

\underline{\CLEFVR} is a challenging new \textit{concept-incremental} NeSy-CL benchmark based on CLEVR~\citep{johnson2017clevr}.
Inputs $\vx$ are renderings of two randomly placed 3D objects with several possible shapes, colors, materials, and sizes.
The goal is to predict whether the objects have the same color, same shape, both, or neither.
The knowledge simply defines the three labels using four (one-hot encoded) concepts encoding shape and color of the two objects.
%
%
There are $5$ tasks.  Objects in each task have only two colors out of ten and two shapes out of ten, with no overlap between tasks.  Knowledge and labels allow for a large number of \textit{reasoning shortcuts}, as illustrated in \cref{fig:main} and detailed in \cref{sec:shortcut-solutions}.
To evaluate quality of learned concepts, we also define an OOD test set containing \textit{unseen combinations} of training objects. 
Overall, the dataset contains almost $5.5$k training data, $500$ data for validation and $2.5$k data for test.

\textbf{Metrics.}  We evaluate all models using common CL metrics~\citep{delange2021continual}, namely class-incremental accuracy (Class-IL) of labels $\vY$ and concepts $\vC$, and forward transfer of labels (FWT).
For \MNISTShortcut and \CLEFVR we also measure label and concept accuracy on the OOD test set.

\textbf{Competitors.}  We compare \method against the following \textit{label-based} continual strategies:
\underline{\naive} fine-tunes the old model on each new task without any continual strategy.
\underline{\restart} fits a model from scratch for each task, without any continual strategy.
\restart and \naive serve as baselines to quantify the impact of forgetting.
\underline{\lwf}: Learning without Forgetting~\citep{li2017learning}, a regularization approach that performs knowledge distillation from the past model.
\underline{\ewc}: Elastic Weight Consolidation~\citep{kirkpatrick2017overcoming}, a regularization approach that avoids drastic updates to important parameters based on the Fisher values of the previous model.
\underline{\er}: Experience Replay~\citep{riemer2019learning}, a popular rehearsal approach that stores a random selection of past examples and replays them when training on the new task.
%
%
\underline{\der}: Dark Experience Replay~\citep{buzzega2020dark}, a state-of-the-art rehearsal approach similar to \er that stores and distills the logits of the past model.
\underline{\derpp}: an improvement of \der that also stores the true label.
We do not compare against prototype-based strategies, like iCaRL \cite{rebuffi2017icarl}, because class prototypes cannot be easily defined in structured representation spaces.
We also consider \underline{\offline} learning over the union of all tasks as an ideal upper bound.
\underline{\method} is implemented as in \cref{eq:cool}.  Following \der, replay examples are selected with \textit{reservoir sampling}~\citep{vitter1985random}, an efficient incremental method for random sampling with uniformity guarantees.
%
All hyperparameters were chosen to optimize last-task Class-IL on the validation labels.
%

\begin{table}[!t]
    \begin{scriptsize}
        \begin{center}
            \input{tables/MNIST-ADD-CBM-vs-NESY}
        \end{center}
    \end{scriptsize}
    \vspace{-1em}
    \caption{\textbf{Knowledge helps, and \method helps even more.}
    Top block: results on \MNISTSeq for all competitors + \CBM with $10\%$ concept supervision, averaged over $10$ seeds.
    Bottom block: same for {\sc DeepProbLog} with $0\%$ supervision.
    \method + DeepProbLog outperforms the neural baseline (despite the gap in supervision) and the other continual strategies.
    The additional results in \cref{sec:additional-results} support these conclusions.
    }
    \label{tab:mnist-seq}
    \vspace{-2em}

\end{table}

\begin{figure*}[!t]
    \centering
    \begin{scriptsize}
    \begin{tabular}{lccc|cc}
        & {\sc Class-IL} ($\vY$) & {\sc Class-IL} ($\vC$) & {\sc OOD Accuracy} ($\vY$) & & \sc Predicted Concepts \\ 
        \rotatebox{90}{\hspace{1em} \MNISTShortcut} &
        \includegraphics[width=0.2\textwidth]{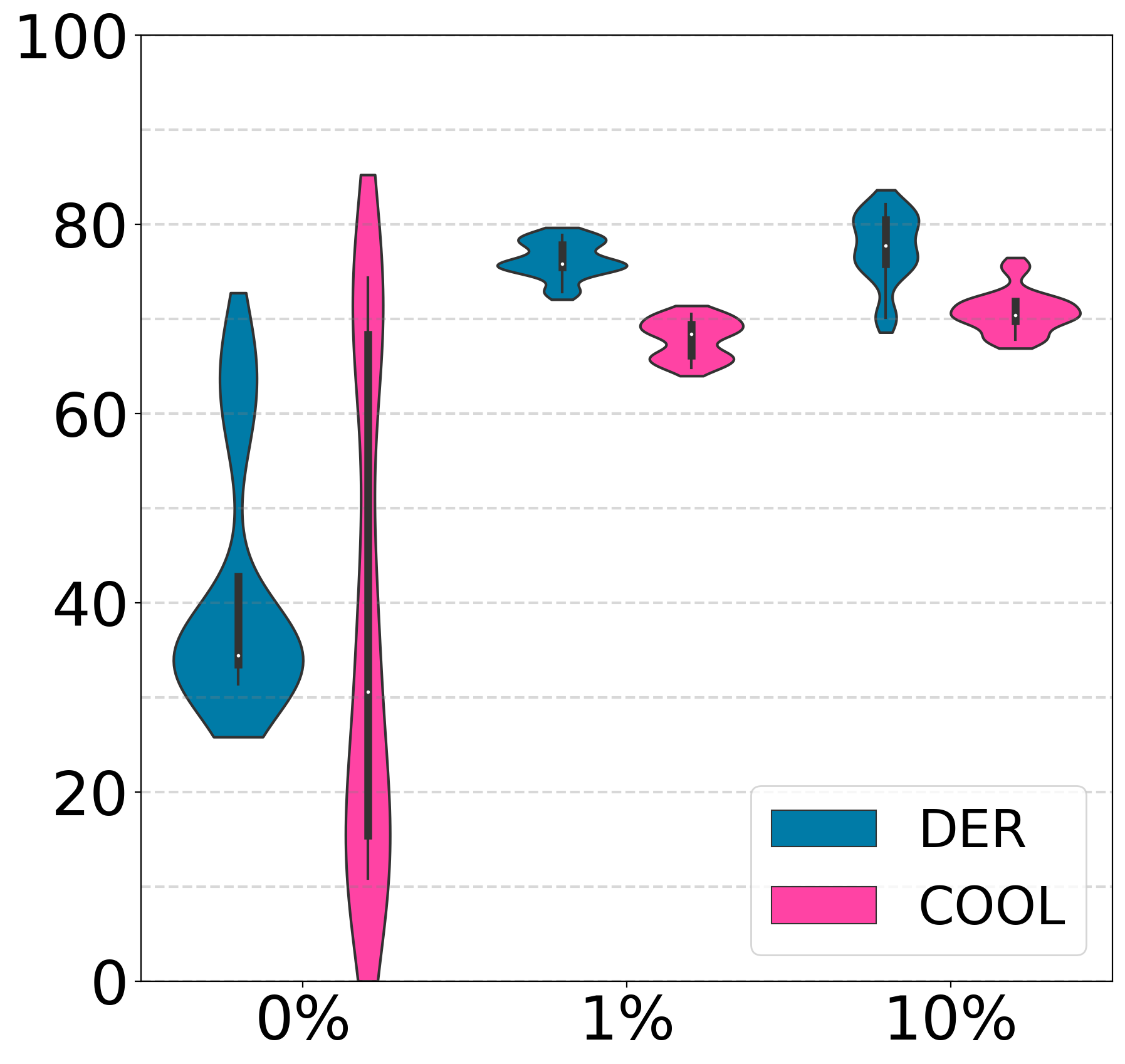} &  
        \includegraphics[width=0.2\textwidth]{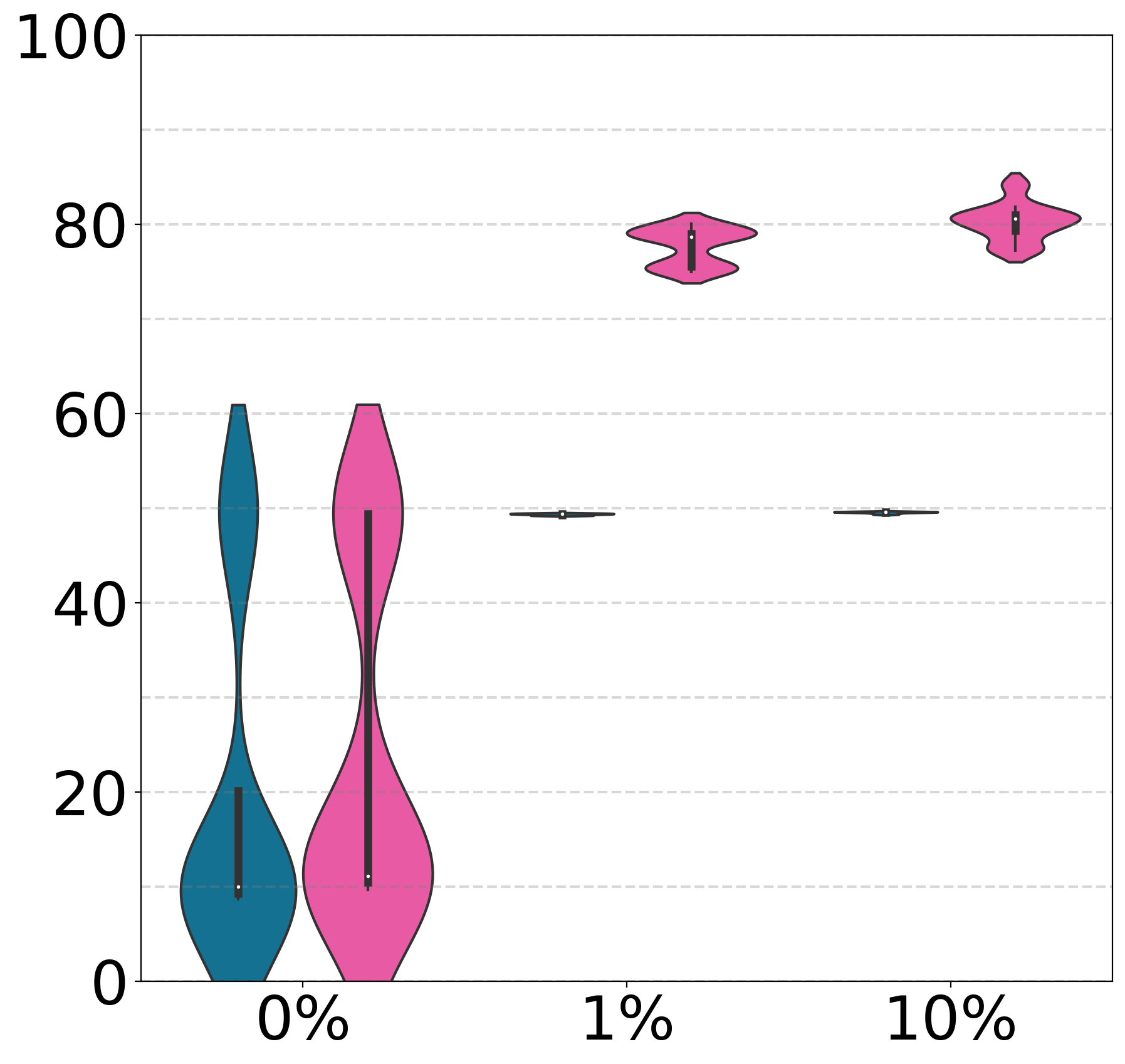} &
        \includegraphics[width=0.2\textwidth]{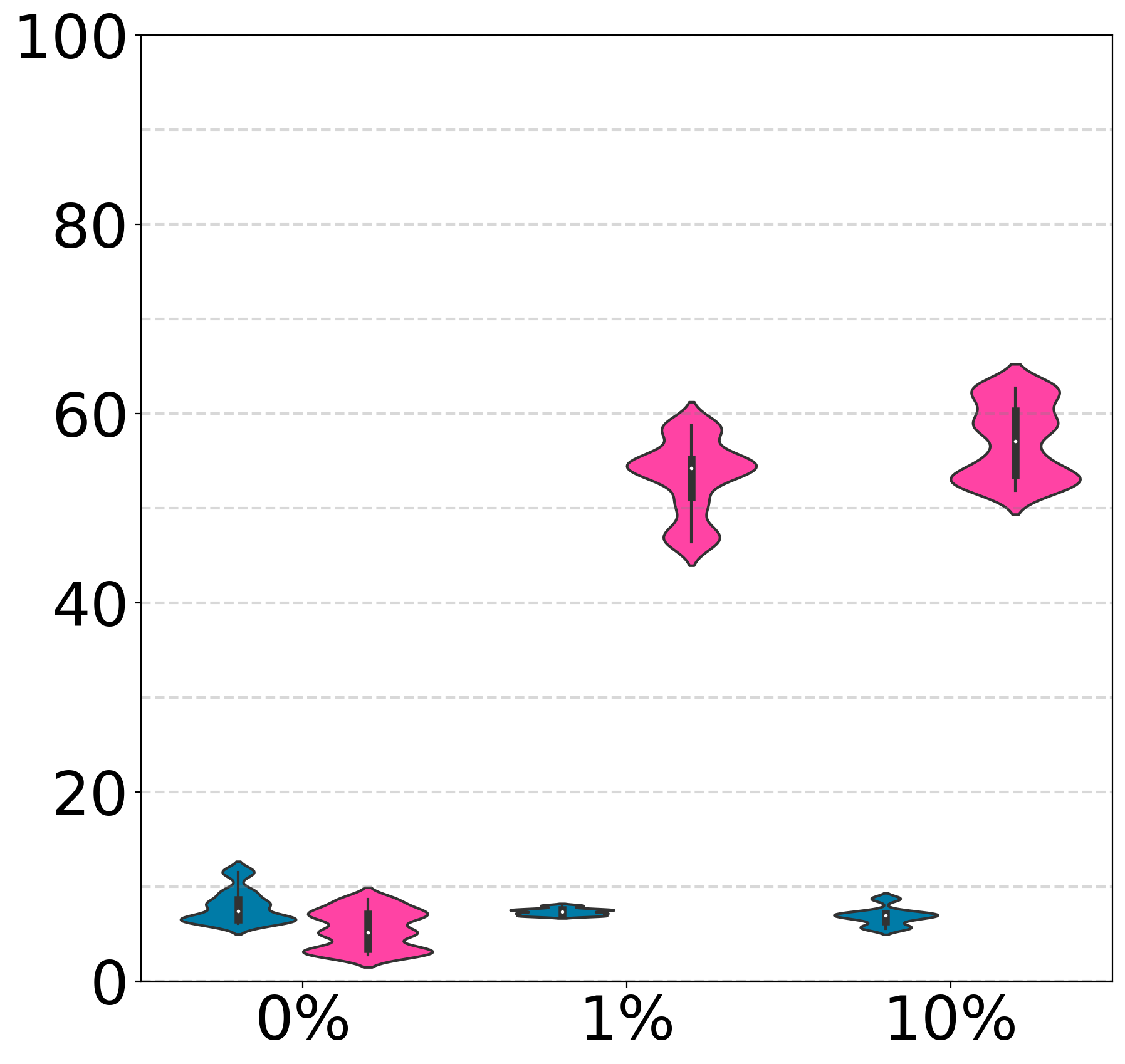} & 
        \rotatebox{90}{\hspace{4em} \sc Ground-Truth} & 
        \includegraphics[width=0.2\textwidth]{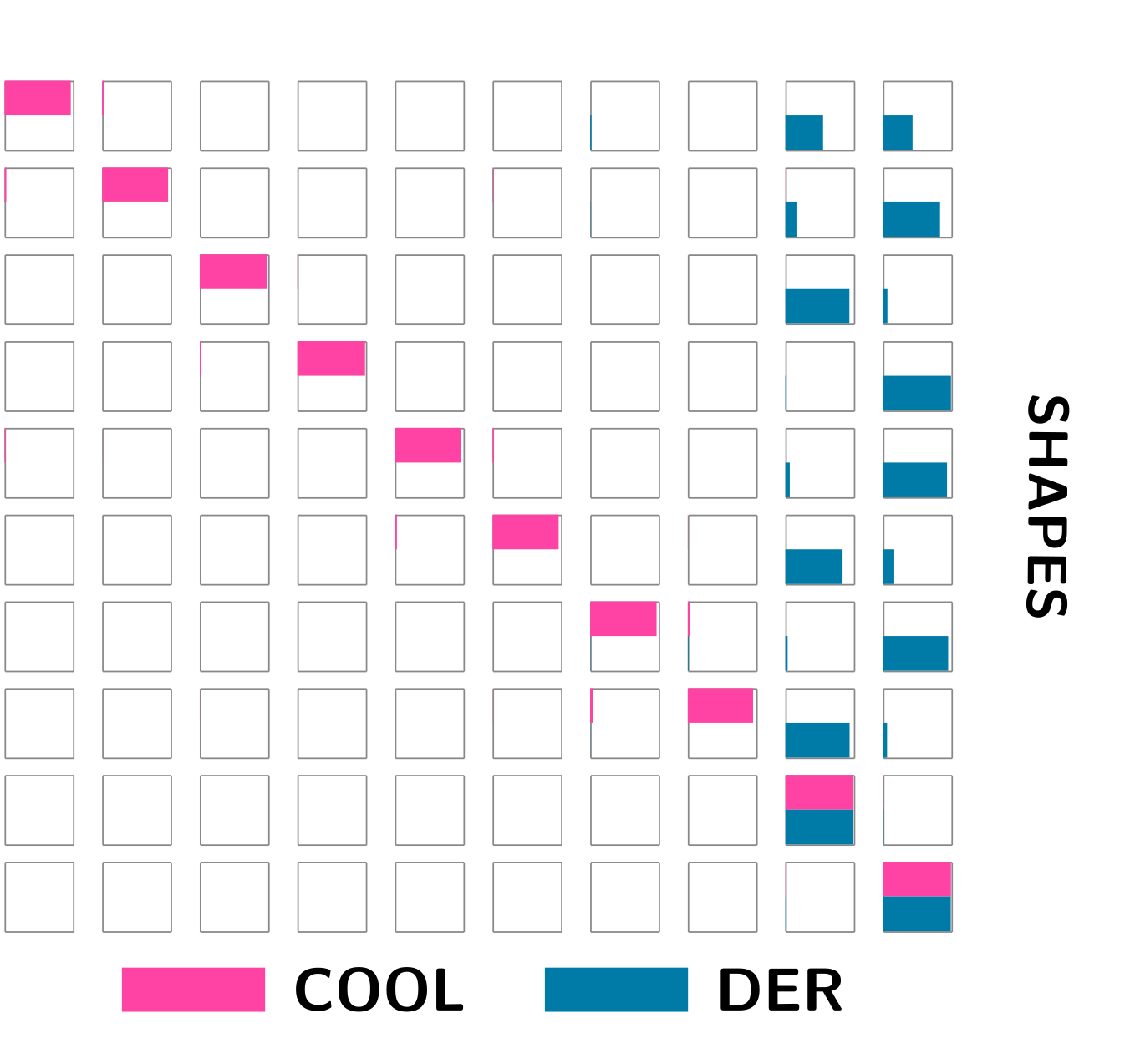} \\
        \rotatebox{90}{\hspace{3em} \CLEFVR} &
        \includegraphics[width=0.2\textwidth]{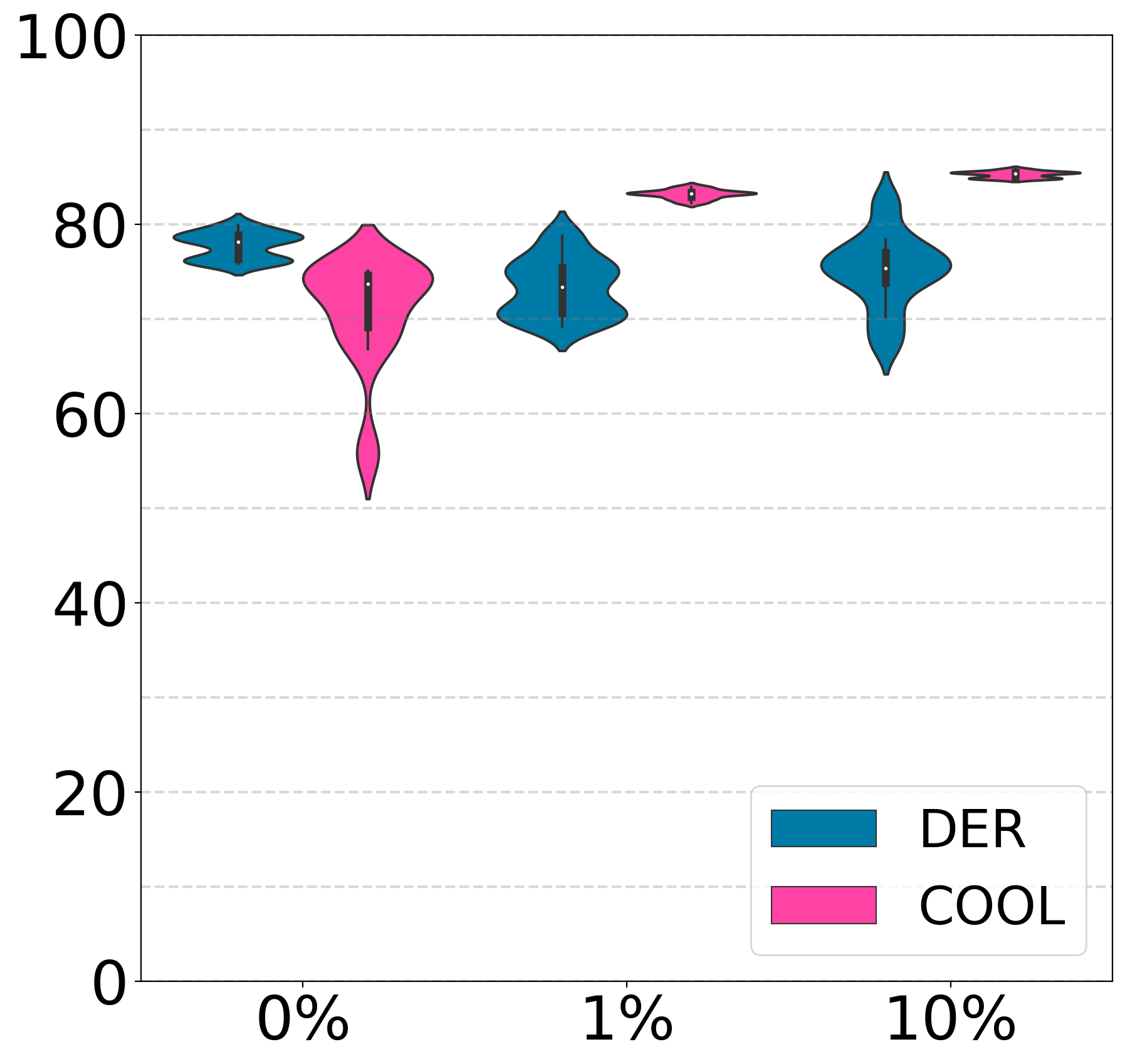} &  
        \includegraphics[width=0.2\textwidth]{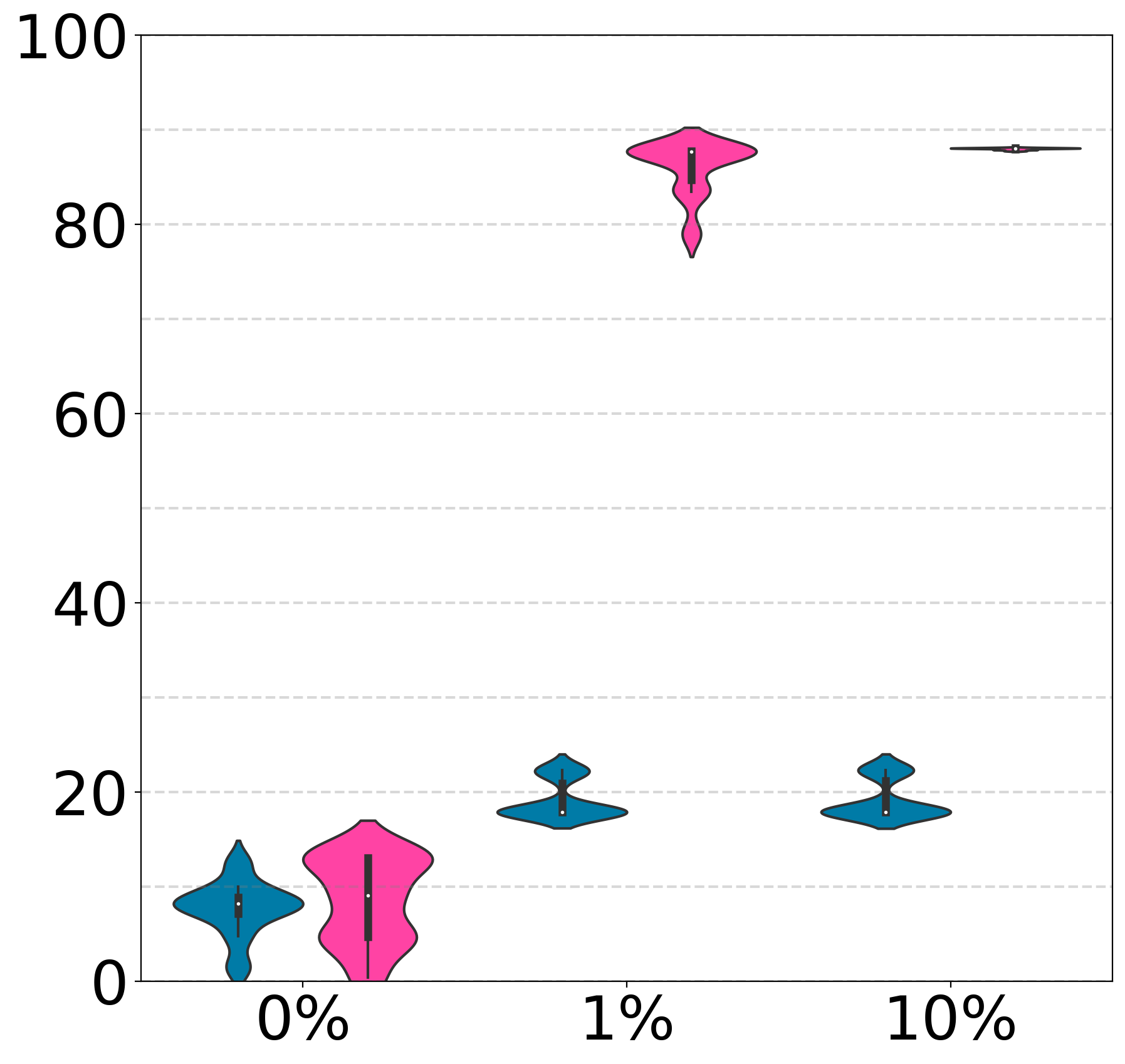} & 
        \includegraphics[width=0.2\textwidth]{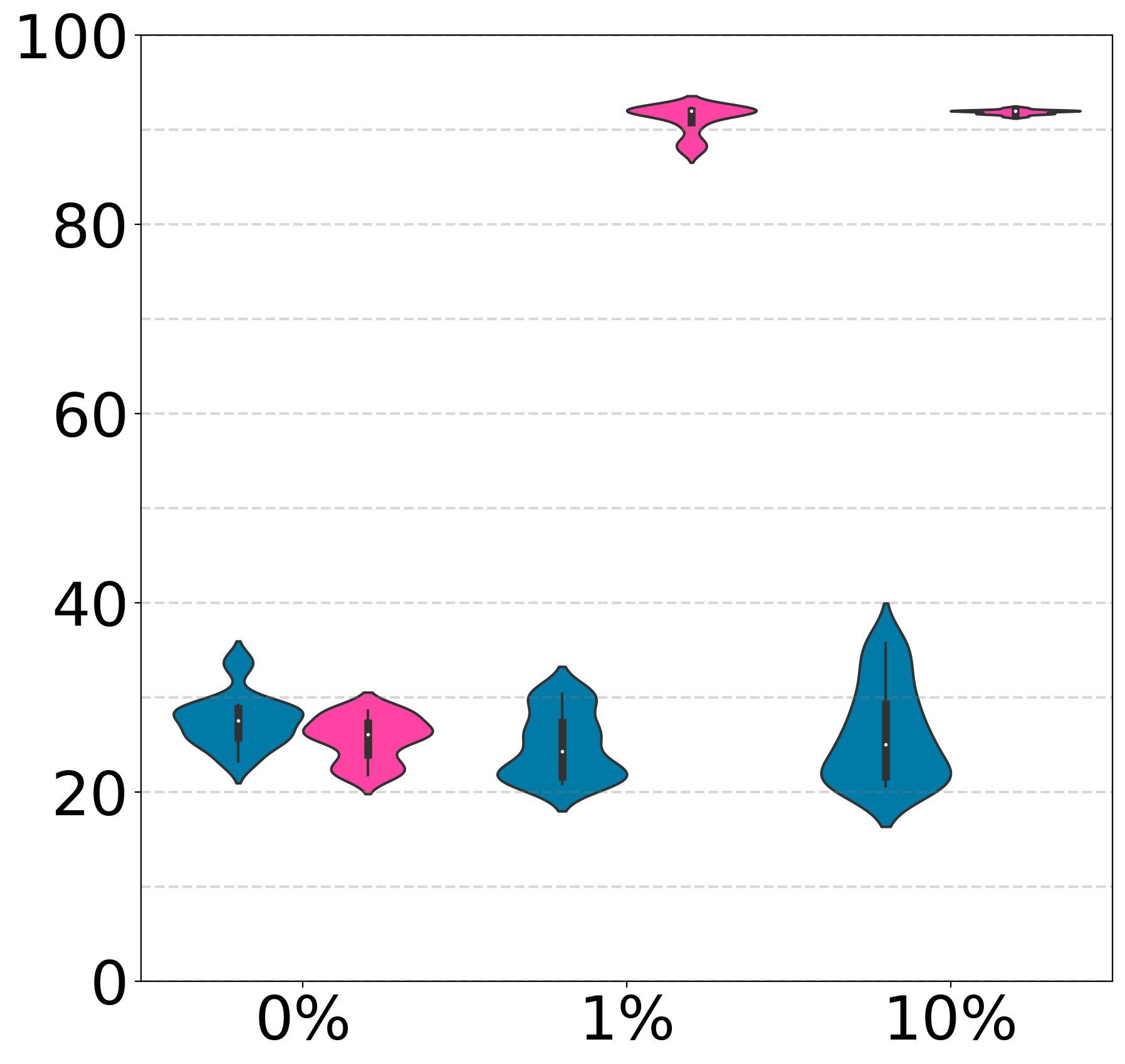} &
        \rotatebox{90}{\hspace{4em} \sc Ground-Truth} &
        \includegraphics[width=0.2\textwidth]{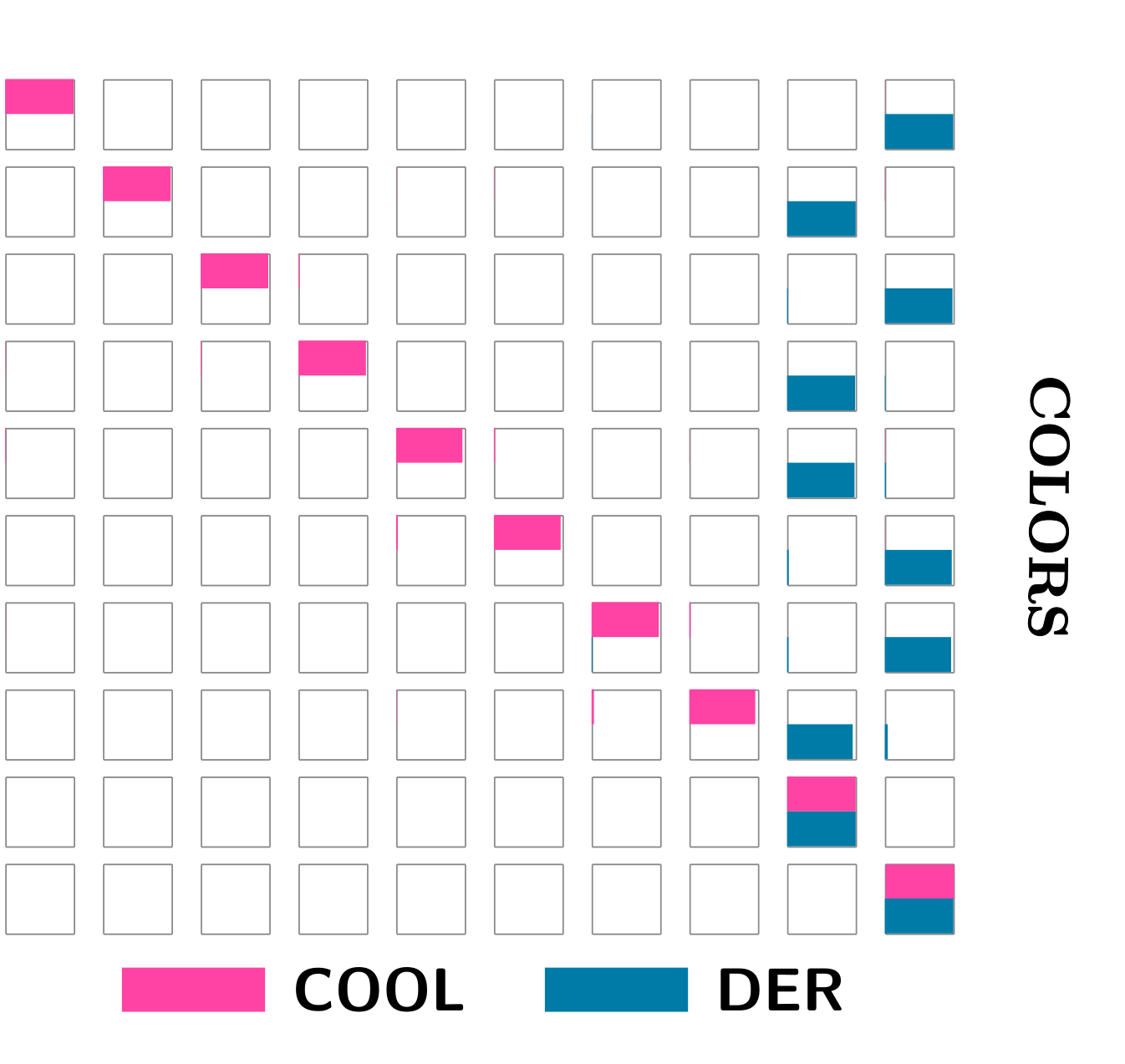}
    \end{tabular}
    \end{scriptsize}
    \vspace{-1em}
    \caption{\textbf{\method avoids shortcuts with few concept-annotated examples.} (Left)  Class-IL (on labels and concepts) and OOD accuracy for \der and \method on \MNISTShortcut (top) and \CLEFVR (bottom). 
    The $x$-axis is the $\%$ of concept annotated examples per task.
    (Right) Confusion matrices of {\tt shape} (top) and {\tt color} (bottom), computed on the last task of \CLEFVR, obtained by DeepProbLog paired with \method (in \textbf{\textcolor{ViolinBlue}{blue}}) vs. \der (in \textbf{\textcolor{ViolinRed}{red}}) with $10\%$ concept supervision.  \method is the only strategy that acquires and maintains the intended semantics. The complete numerical results are reported in \cref{sec:additional-results}. 
    }
    \label{fig:q2-q3}
    \vspace{-1.3em}
\end{figure*}

\textbf{Q1: Knowledge helps, \method helps even more.}  We evaluate the impact of knowledge on \MNISTSeq, where shortcuts are \textit{absent}.  Specifically, we evaluate different continual strategies paired with DeepProbLog~\citep{manhaeve2018deepproblog} and Concept-bottleneck Models (\CBMs)~\citep{koh2020concept}.
Both architectures extract concepts using a convolutional network, but differ in how they infer the label.  DeepProbLog uses a probabilistic-logic layer encoding the available knowledge (see \cref{sec:deepproblog}).  \CBMs aggregate the concepts using a learnable network \textit{independent} of the knowledge, and serve as a purely neural baseline.
Additional architectural choices are reported in \cref{sec:implementation-details}.
The two models were trained for $25$ epochs per task, using a fixed buffer size of $1000$, but received different amounts of concept supervision:  $10\%$ for \CBMs, which is enough to learn the intended concepts in the offline setting, and \textit{none} for DeepProbLog.

The \textit{offline} performance of \CBM and DeepProbLog are excellent, achieving around $96\%$ label accuracy and $99\%$ concept accuracy, showing that both are capable of solving the learning task.
In the \textit{continual} setting, however, the gap between the neural and NeSy models widens noticeably.  The results are reported in Table \ref{tab:mnist-seq}.  Despite DeepProbLog being harder to learn than \CBMs (as shown by \restart and \naive), all replay strategies -- \ie \er, \der, \derpp, and \method -- perform much better when paired with DeepProbLog than \CBMs:  remarkably, label Class-IL sees gains close to $50\%$ for \der, and similarly FWT for \method.  This highlights the benefits of knowledge, which apply despite DeepProbLog having access to \textit{no concept supervision}.
The regularization strategies \lwf and \ewc are not informative, as they struggle to improve on the \naive and \restart baselines both \textit{with} and \textit{without} knowledge.\footnote{The only exception is \lwf on \CBM, which displayed pathological behavior, see \cref{sec:additional-results}.}

Overall, \cref{tab:mnist-seq} indicates that \textit{when reasoning shortcuts are absent}, knowledge facilitates identifying better concepts, and thus better predictions.  By retaining these concepts, \method manages to outperform all other competitors on both \CBMs and DeepProbLog.
The runner-up, \der, keeps up only when knowledge is available and with a substantial margin in terms of FWT ($45\%$ vs. $83\%$).
In contrast, even though \CBMs can acquire good concepts, this does not always yield good predictions, chiefly because the top layer undergoes forgetting, cf. \cref{sec:knowledge-helps}.  Thanks to knowledge, DeepProbLog avoids this issue altogether.

\textbf{Q2: \method avoids reasoning shortcuts.}  Next, we evaluate the impact of concept quality and rehearsal in \MNISTShortcut and \CLEFVR, which \textit{are} affected by reasoning shortcuts.
Given the sub-par performance of \CBMs, we focus on DeepProbLog from now on.
Also, we restrict our attention to \method and \der, the runner up in the previous experiment.  Results for all other competitors are available in \cref{sec:additional-results}.
For \MNISTShortcut we set a buffer size of $1000$ and $100$ epochs per task, and to $250$ examples and $50$ epochs for \CLEFVR.

The results in \cref{fig:q2-q3} shows that, when no concept supervision is in place, the presence of shortcuts complicates retaining the correct concepts, as displayed by low values of Class-IL $(\vC)$.  This does not impact directly Class-IL $(\vY)$ in the case of \CLEFVR, but yields extremely low OOD generalization, around $10\%$ for \MNISTShortcut and $25\%$ for \CLEFVR.
The effect on increasing concept supervision on \der is only seemingly positive, as label accuracy does improve in both data sets.  However, Class-IL on concepts (about $20$--$50\%$) and OOD accuracy ($10\%$--$30\%$) are very poor, despite the supervision.  What happens is precisely the issue depicted in \cref{fig:main}:  the model acquires good concepts for one task, but -- due to reasoning shortcuts and lack of concept rehearsal -- these get corrupted when fitting on the next task.
Since \method retains the high quality concepts identified via supervision, the latter leads to clear improvements in label accuracy, concept accuracy \textit{and} OOD accuracy for \method.  As a result, \method improves on \der by about $+30\%$ and $+60\%$ in terms of Class-IL ($\vC$) and $+40\%$ and $+60\%$ in OOD, in the two data sets respectively.

We stress that label-based strategies inevitably fall for reasoning shortcuts \textit{even if concept supervision is provided}.  This is clearly shown by the concept confusion matrices reported in \cref{fig:q2-q3} (right).  Notice that, out of \textit{all} strategies, only \method manages to prevent shortcuts.  Further details are available in \cref{sec:additional-results}.


\textbf{Q3: \method requires minimal concept supervision.}  Figure \ref{fig:q2-q3} 
shows that \method identifies high-quality concepts when given dense annotations for only $1\%$ of the training set.  This translates to about $30$ examples per task in \MNISTShortcut, and to only $12$ in \CLEFVR.
Increasing concept supervision to $10\%$ improves Class-IL ($\vC$) by $3\%$ and shrinks its variance, but $1\%$ is enough to substantially outperform \der in our tests.

\section{Related Work}
\label{sec:related-work}

\noindent
\textbf{Neuro-symbolic integration.}  NeSy encompasses a diverse family of methods integrating learning and reasoning~\citep{de2021statistical}.  Here, we focus on approaches for encouraging neural networks to output structured predictions consistent with prior knowledge.
The two main strategies introduce an additional loss penalizing inconsistent predictions~\citep{xu2018semantic,fischer2019dl2,ahmed2022neuro} or a top reasoning layer~\citep{manhaeve2018deepproblog,giunchiglia2020coherent,hoernle2022multiplexnet,ahmed2022semantic}.  Since the former cannot guarantee that the model outputs consistent predictions, we focus on the latter.
In either case, end-to-end training requires to differentiate through the knowledge.  One option is to soften the knowledge using fuzzy logic~\citep{diligenti2012bridging,donadello2017logic}, but doing so can introduce semantic and learning artifacts~\citep{giannini2018convex,van2022analyzing}.
An alternative is to cast reasoning in terms of probabilistic logics~\citep{de2015probabilistic}, which preserves semantics and allows for sound inference and learning.
\changed{DeepProbLog is just an example of Nesy strategies} 
~\citep{manhaeve2021approximate,huang2021scallop,winters2022deepstochlog,ahmed2022semantic,van2022anesi}.
All NeSy approaches are \textit{offline} and suffer from catastrophic forgetting, and existing continual strategies do not protect them from reasoning shortcuts, as shown in \cref{sec:experiments}.
Since these depend only on the latent nature of concepts, they affect probabilistic-logic and fuzzy logic architectures alike.  \method applies to all these, \changed{cf. \cref{sec:applicability}}.

\textbf{Continual Learning.}  CL algorithms attempt to preserve model plasticity while mitigating catastrophic forgetting~\citep{robins1995catastrophic} using a variety of techniques~\citep{van2022three,qu2021recent}.
%
A first group of strategies, like Experience Replay~\citep{riemer2019learning} and {\sc er-ace}~\citep{caccia2021new}, store and rehearse a limited amount of examples from previous tasks.
Doing so ignores additional ``dark knowledge'' learned by the past model, so techniques like \der~\citep{buzzega2020dark}, \derpp, and others~\citep{rebuffi2017icarl,li2017learning,castro2018end,hou2019learning},
drop rehearsal in favor of distillation.  \method follows the same strategy.
%
%
Popular alternatives include architectural approaches~\citep{rusu2016progressive}, which freeze or add model parameters as needed, and regularization strategies~\citep{delange2021cpe,kirkpatrick2017overcoming,Aljundi_2018_MAS,zenke2017SI}.
These introduce extra penalties in the loss function to discourage changing parameters essential for discriminating classes, but can struggle with complex data~\citep{aljundi2019gradient}.
To the best of our knowledge, CL has only been tackled in flat prediction settings (e.g., classification), and existing strategies focus on preserving label accuracy only.
The only work on forgetting in CBMs is~\citep{marconato2022catastrophic}, which however ignores knowledge altogether.

\textbf{Reasoning shortcuts.}  In machine learning, ``shortcuts'' refer to models that exploit \textit{spurious} correlations between inputs and annotations to achieve high training accuracy~\citep{ross2017right,lapuschkin2019unmasking}.
%
Proposed solutions include dense annotations~\citep{ross2017right}, out-of-domain data~\citep{parascandolo2020learning}, and interaction with annotators~\citep{teso2022leveraging}.
\citet{stammer2021right} have investigated shortcuts in NeSy and proposed to fix them using knowledge, under the assumption that concepts are high-quality.  We make no such assumption.
Our work is the first to investigate \textit{reasoning} shortcuts that knowledge cannot always prevent and their preminence in NeSy-CL.

\section{Conclusion}

We initiated the study of Neuro-Symbolic Continual Learning and showed that knowledge, although useful, can be insufficient to prevent acquiring reasoning shortcuts that compromise concept semantics and cross-task transfer.
Our approach, \method, acquires and preserves high-quality concepts, attaining better concepts and performance than existing CL strategies in three new NeSy-CL benchmarks.

 \section*{Acknowledgements}
 \changed{
 We acknowledge the support  of the MUR PNRR project FAIR - Future AI Research (PE00000013) funded by the NextGenerationEU. The research of AP and ST was partially supported by TAILOR, a project funded by EU Horizon 2020 research and innovation program under GA No 952215. We acknowledge the CINECA award under the ISCRA initiative, for the availability of high performance computing resources and support. The research of SC was partially supported by Italian Ministerial grant PRIN 2020 “LEGO.AI: LEarning the Geometry of knOwledge in AI systems”, n. 2020TA3K9N. The research of EF was partially supported by the European Union’s Horizon 2020 research and innovation program DECIDER under Grant Agreement 965193. We acknowledge Angelo Porrello for his useful discussion with us. }

\bibliographystyle{icml2022}
\bibliography{explanatory-supervision,references}

\newpage
\appendix
\onecolumn


\section{Proofs}
\label{sec:proofs}

\subsection{Proof of \cref{thm:ll-optima-satisfy-bk}}

Taking \cref{eq:deepproblog-joint} as reference, the log-likelihood of $\dataset$ can be rewritten as:
\[
    \sum_{(\vx, \vy) \in \dataset} \log p_\theta(\vy \mid \vx; \BK)
    = \sum_{(\vx, \vy) \in \dataset} \log \ \inner{ u_{\BK}( \vy \mid \cdot) }{ p_\theta( \cdot \mid \vx ) }
    \label{eq:ll-all-knowledge}
\]
Here, the inner product runs over all possible values of $\vc$.
To see what the optima of this quantity look like, fix a single example $(\vx, \vy) \in \dataset$ and let $\calC_{\vy}$ be the set of values $\vc$ that satisfy the knowledge $\BK[\vY/\vy]$ and $\calC_{\bar{\vy}}$ be those that violate the knowledge.
The likelihood of $(\vx, \vy)$ amounts to:
\[
    \inner{ u_{\BK}( \vy \mid \cdot) }{ p_\theta( \cdot \mid \vx ) }
        = \sum_{\vc \in \calC_{\vy}} u_{\BK}( \vy \mid \vc) p_\theta( \vc \mid \vx )
        + \sum_{\vc \in \calC_{\bar{\vy}}} \underbrace{u_{\BK}( \vy \mid \vc)}_{=0} p_\theta( \vc \mid \vx )
    \label{eq:inconsistent-concepts-do-not-affect-the-likelihood}
\]
The inner product is maximized whenever $p_\theta(\vC \mid \vx)$ allocates \textit{all} probability mass to values $\vc$ that satisfy $\BK[\vY/\vy]$, because the remaining ones do not contribute anything to the likelihood.
Hence, in order to maximize \cref{eq:ll-all-knowledge} it is sufficient that, for every $(\vx, \vy) \in \dataset$, $p_\theta(\vC \mid \vx)$ assigns zero probability to all concept configurations $\vc$ that are ruled out by the knowledge $\BK[\vY/\vy]$.
\begin{flushright}
$\square$
\end{flushright}

\textbf{Remarks:}  This result essentially states that DeepProbLog's reasoning layer has the same effect as the Semantic Loss~\citep{xu2018semantic} on the underlying neural network, and it is of independent interest.
Notice that \cref{thm:ll-optima-satisfy-bk} also holds for non-uniform label distributions without any change to the proof.

\subsection{Proof of \cref{thm:concept-distillation-is-enough}}

We start by proving a general lemma:

\begin{lemma}
\label{lemma:concept-distillation-is-enough}
Consider tasks $\task^{1}, \ldots, \task^{(t)}$ and two parameter configurations $\varphi, \psi \in \Theta$.  If both models assign non-zero likelihood to all examples in $\dataset^{1:t}$, there exists a finite constant $\gamma$, depending only on the model architecture, knowledge and data, such that:
\[
    |\calL(\varphi, \task^{1:t}) - \calL(\psi, \task^{1:t})|
    \le
    \gamma \sum_{s \le t} \sum_{(\vx, \vy) \in \dataset^{(s)}} \norm{
        p_\varphi(\vC \mid \vx) - p_\psi(\vC \mid \vx)
    }_1
\]
\begin{proof}
The left-hand side can be expanded to:
\begin{align}
    & \frac{1}{t} \left|
        \sum_{s \le t} \frac{1}{|\dataset^{(s)}|} \sum_{(\vx, \vy) \in \dataset^{(s)}} \left(
            \log p_\varphi(\vy \mid \vx; \BK^{(s)}) - \log p_\psi(\vy \mid \vx; \BK^{(s)})
        \right)
    \right|
    \\
    & \quad \le \frac{1}{t}
        \sum_{s \le t} \frac{1}{|\dataset^{(s)}|} \sum_{(\vx, \vy) \in \dataset^{(s)}} \left|
            \log p_\varphi(\vy \mid \vx; \BK^{(s)}) - \log p_\psi(\vy \mid \vx; \BK^{(s)})
        \right|
    \label{eq:lldiff}
\end{align}
Recall that, for any $a, b \in [\beta, \infty]$, it holds that $|\log a - \log b| \le \frac{1}{\beta} |a - b|$.
For any $(\vx, \vy)$ and task $s \le t$, it holds that:
\begin{align}
    |\log p_\varphi(\vy \mid \vx; \BK^{(s)}) - \log p_\psi(\vy \mid \vx; \BK^{(s)})|
        & \le \frac{1}{\beta} \big| p_\varphi(\vy \mid \vx; \BK^{(s)}) - p_\psi(\vy \mid \vx; \BK^{(s)}) \big|
    \\
    = \frac{1}{\beta} \big| \inner{ u_{\BK^{(s)}}(\vy \mid \cdot) }{ p_\varphi(\cdot \mid \vx) - p_\psi(\cdot \mid \vx) } \big|
        & \le \frac{1}{\beta} \norm{ u_{\BK^{(s)}}(\vy \mid \cdot) }_\infty \cdot \norm{ p_\varphi(\cdot \mid \vx) - p_\psi(\cdot \mid \vx) }_1
\end{align}
The first step follows by choosing
$
    \beta := \min_{(\vx, \vy) \in \dataset^{1:t}} \min_{\theta \in \{ \varphi, \psi \}} p_\theta(\vy \mid \vx; \BK^{(s)}) > 0
$,
the second one from \cref{eq:deepproblog-joint}, and the last one from H\"older's inequality.
By \cref{eq:deepproblog-rules}, the max norm of $u_{\BK^{(s)}}$ amounts to:
\[
    \norm{ u_{\BK^{(s)}}(\vy \mid \cdot) }_\infty
    = \max_{\vc} \frac{ \Ind{(\vc, \vy) \models \BK^{(s)}} }{ Z(\vc; \BK^{(s)}) }
    = \frac{1}{\min_{\vc} Z(\vc; \BK^{(s)})}
    \le \frac{1}{\zeta}
    \le 1
\]
where we chose $\zeta = \min_{\vc} \min_{s \le t} Z(\vc; \BK^{(s)})$.  Therefore, 
\[
    \frac{1}{\beta} \norm{ u_{\BK^{(s)}}(\vy \mid \cdot) }_\infty \cdot \norm{ p_\varphi(\cdot \mid \vx) - p_\psi(\cdot \mid \vx) }_1
    \le \frac{1}{\beta \zeta} \norm{ p_\varphi(\cdot \mid \vx) - p_\psi(\cdot \mid \vx) }_1
\]
Taking $\gamma = \max_s \frac{1}{\beta \zeta |\dataset^{(s)}| t}$ and replacing the above in \cref{eq:lldiff} yields the claim.
\end{proof}
\end{lemma}

In order to prove \cref{thm:concept-distillation-is-enough}, let $\theta^{(t-1)}$ be the parameters learned after observing $t - 1$ tasks and $\theta$ to be learned at the current iteration.
Applying \cref{lemma:concept-distillation-is-enough} with $\varphi = \theta^{(t-1)}$ and $\psi = \theta$ to \cref{eq:average-risk} yields the desired result.

\textbf{Remark:}  In the worst case $\zeta$ can be as small as $1$, which occurs if in all tasks $s \le t$ the knowledge $\BK^{(s)}$ accepts a single concept configuration $\vc$ for every example $(\vx, \vy)$;  more commonly, $\zeta$ is exponential in the number of concepts $k$.
Also, $\beta$, which is the minimum likelihood attained by either $p_\varphi$ or $p_\psi$ on the data sets $\dataset^{(1)}, \ldots, \dataset^{(t)}$, is actively maximized during learning.

\subsection{What Are Correct Semantics?}
\label{sec:semantics}

Strictly speaking, the only concept distribution with the actual \textit{correct semantics} is the ground-truth distribution $p(\vC \mid \vx)$.

Our assumption is that the ground-truth concept distribution we are given is always consistent with the knowledge, in the sense that $p(\vC \mid \vx) \models \BK^{(t)}[\vX/\vx, \vY/\vy]$ for every possible task $\task^{(t)} = (\dataset^{(t)}, \BK^{(t)})$ and example $(\vx, \vy) \in \dataset^{(t)}$.  In other words, we assume that the knowledge correctly reflects how the world works.
Under this assumption, having the correct semantics is useful \textit{in practice} because, when paired with the knowledge, the ground-truth distribution by construction always yields correct labels in every possible task the learner can in principle receive.  This is a form of \textit{systematic generalization}.

Naturally, if the learned distribution $p_\theta(\vC \mid \vx)$ matches the ground-truth distribution exactly, then it will also achieve systematic generalization.  This condition, however, is very restrictive.
Pragmatically, we can relax this requirement, and say that a distribution encodes the correct semantics if it is \textit{indistinguishable from the ground-truth distribution in terms of what concepts it predicts}.
Formally, we say that a distribution $p_\theta(\vC \mid \vx)$ is \textit{semantically equivalent} to the ground-truth distribution on data $\dataset$ if it allows us to infer the same concept configuration for all data points, or formally:
\[
    \forall \vx \in \dataset \ . \ \argmax_{\vc} \ p(\vc \mid \vx) \equiv \argmax_{\vc} \ p_\theta(\vc \mid \vx)
    \label{eq:same-semantics}
\]
where we used $\equiv$ to indicate set equivalence.
This is quite intuitive:  any distribution satisfying \cref{eq:same-semantics} will yield the same concepts $\vc$ as $p(\vC \mid \vx)$ for every $\vx$, and therefore also the same MAP states $\vy$ under any choice of knowledge $\BK$.

The opposite is \textit{not} generally true:  the knowledge $\BK$ might have multiple possible solutions, in the sense that different choices of concepts $\vc$ might yield the same label $\vy$.  In this case, the label does not carry enough information to recover the ground-truth concepts $\argmax_{\vc} p(\vc \mid \vx)$, and therefore also a concept distribution $p_\theta(\vC \mid \vx)$ semantically equivalent to $p(\vC \mid \vx)$.  This is exactly what we mean by \textit{reasoning shortcuts}: concept distributions that achieve high performance on the observed task(s) but have no guarantee of systematically generalizing to future tasks, or more formally:
\[
    \argmax_{\vy} p(\vy \mid \vx; \BK) \equiv \argmax_{\vy} p_\theta(\vy \mid \vx; \BK)
    \qquad \land \qquad
    \argmax_{\vc} p(\vc \mid \vx) \not\equiv \argmax_{\vc} p_\theta(\vc \mid \vx)
\]
%

\subsection{Reasoning Shortcuts in other NeSy Architectures}
\label{sec:applicability}

\changed{
While \cref{thm:ll-optima-satisfy-bk} shows that reasoning shortcuts do affect DeepProbLog, which maximizes for label likelihood, we remark that they are a \textit{general} phenomenon.
It is easy to see that reasoning shortcuts occur whenever the prior knowledge admits deducing the correct \textit{label} $\vy$ from \textit{concepts} $\vc$ that do not have the correct semantics, as this makes it is impossible for a model to distinguish between concepts with ``correct'' vs.\@ ``incorrect'' semantics by maximizing accuracy alone.
This impacts \textit{offline} NeSy prediction tasks and NeSy-CL problems alike;  indeed, \cref{thm:ll-optima-satisfy-bk} makes no assumption on how the training set $\dataset$ has been generated.
}

\changed{
Reasoning shortcuts are not specific to DeepProbLog.  On the contrary, this situation can be triggered by a variety of other NeSy architectures, including but not limited to:}
\begin{itemize}[leftmargin=1.25em]

    \item[(\textit{i})] \changed{NeSy predictors that rely on a top reasoning layer to ensure predictions are consistent with prior knowledge, which are typically trained to maximize some surrogate of the label accuracy, including~\citep{ahmed2022semantic,giunchiglia2020coherent,hoernle2022multiplexnet,huang2021scallop,winters2022deepstochlog,van2022anesi}.}

    \item[(\textit{ii})] \changed{NeSy predictors that rely on \textit{relaxed} reasoning layers obtained by softening the logical prior knowledge~\citep{diligenti2012bridging,donadello2017logic}, because this transformation usually preserves existing optima of the label accuracy.  As such, it also preserves unintended optima -- that is, reasoning shortcuts.}

    \item[(\textit{iii})] \changed{Neural networks trained to maximize accuracy and consistency with prior knowledge using the Semantic Loss and similar techniques~\citep{xu2018semantic,fischer2019dl2,ahmed2022neuro}.  In fact, \cref{thm:ll-optima-satisfy-bk} shows that DeepProbLog is affected by shortcuts precisely because, from the neural network's perspective, its reasoning layer acts exactly like the Semantic Loss; see our remark in \cref{sec:proofs}.}

\end{itemize}

\changed{More generally, reasoning shortcuts impact NeSy tasks and architectures beyond these, at least as long as models are trained by optimizing loss functions that do not measure or correlate with concept quality.  We leave a detailed analysis of specific cases to future work.}

\section{Implementation Details}
\label{sec:implementation-details}

In this Section, we report useful details for the models and the metrics adopted in the evaluation. 

\subsection{Hardware and Software Implementation}

The code for the project was developed on top {\tt mammoth} \citep{boschini2022class}, a well-known CL framework. We included the implementation of \DeepProbLog for \MNISTAdd from \textsc{vael} \citep{misino2022vael}. The generation of \CLEFVR was adapted from \citep{stammer2021right}. All experiments were implemented using Python 3 and Pytorch~\citep{paszke2019pytorch} and run on a server with 128 CPUs, 1TiB RAM, and 8 A100 GPUs.

\subsection{Metrics}
\label{sec:metrics}

We adopted standard CL measures, namely task-incremental (Task-IL) and class-incremental (Class-IL) accuracy, applied here to both labels and concepts predictions, as well as forward transfer (FWT) and backward transfer (BWT) on the labels, see also \citep{buzzega2020dark}.  Below we write $T$ to indicate the last task.
\begin{itemize}

    \item \textbf{Class-IL} measures the average accuracy on the test sets of all tasks $T$.  In \cref{tab:mnist-seq}, we report Class-IL at the very last task $T$, defined as:
    \[
        \textsc{Class-IL}_{\vY}(\theta_T) = \frac{1}{T} \sum_{s=1}^T {\cal A}_{\vY}(\theta_T, s)
        \label{eq:class-il-definition}
    \]
    where ${\cal A}_{\vY}(\theta_t,s )$ denotes the accuracy on the labels evaluated on the test set of task $s$.  Class-IL for concepts is analogous, but builds on the average accuracy over \textit{all} concepts.

    \item \textbf{Task-IL} is the average accuracy over the test sets of all tasks up to $t$, evaluated only on examples annotated with the classes or concepts appearing in task $t$.
    The definition is identical to \cref{eq:class-il-definition} except that we mask the prediction of model so as to place mass only on the labels appearing in $\dataset^s$, with $s \leq t$.

    \item  \textbf{FWT} evaluates the adaptability of the model at each time-step to the successive task.  Formally, at each $t$, FWT measures the average gain in accuracy between $\theta_t$ and a random baseline $\theta_\text{rand}$ when predicting the labels of the task $t+1$.  This can be written as:
    \[
        \label{eq:fwt-definition}
        \textsc{FWT} = \frac{1}{T - 1} \sum_{t=1}^{T-1}   {\cal A}_{\vY}(\theta_t, t+1) - {\cal A}_{\vY}(\theta_\text{rand}, t+1) 
    \]
    where $\theta_\text{rand}$ denotes the initialized model with random weights. 
    
    \item \textbf{BWT} measures how much forgetting the model undergoes by looking at how much accuracy for each task $t$ is retained after the last task.  Formally:
    \begin{equation}
        \textsc{BWT} = \frac{1}{T} \sum_{t=1}^T {\cal A}_{\vY}(\theta_t, t) - {\cal A}_{\vY}(\theta_T, t)
    \end{equation}

\end{itemize}
For the sake of brevity, in the main paper we reported only Class-IL on labels and concepts, and FWT.  Task-IL was omitted because it does not account for accuracy on past concepts that no longer occur in the last task, and BWT because it does not as informative as Class-IL.  All results for these extra metrics on \MNISTSeq are reported in \cref{sec:additional-results}.

\subsection{Architectures \& Models Details}

\underline{\MNISTAdd}:  For both \CBMs and \DeepProbLog we adopted the same architecture for extracting concepts -- henceforth, \textit{encoder} -- and we implemented it as a standard convolutional neural network, with dropout set at $50\%$ after each convolution module. We also inject a noise term $\epsilon \sim  {\cal N} (0, 0.1) $ after the encoder to stabilize the overall training process.  The complete structure is reported in \cref{tab: mnist-addition-backbone}.

\begin{table}[h]
    \centering
    \footnotesize
    \caption{CNN Encoder for \MNISTAdd}
    \begin{tabular}{cccc}
         \toprule
         \textsc{Input shape} & \textsc{Layer type} & \textsc{Parameters} & \textsc{Activation} \\ 
         \midrule
         $(28, 28, 1 )$ & Convolution & {depth}=$64$,  {kernel}=$4$, {stride}=$2$, {padding}=$1$ & ReLU \\
         $(14,14,64)$ & Dropout & $p=0.5$ & \\
         $(14,14, 64)$  & Convolution & depth=$128$,  kernel=$4$, stride=$2$, padding=$1$ & ReLU \\
         $(7,7,128)$ & Dropout & $p=0.5$ & \\
         $(7, 7, 128)$  & Convolution & depth=$256$,  kernel=$4$, stride=$2$, padding=$1$ & ReLU \\
         $(3,3,256) $ & Flatten &   &  \\
         $(2304)$ & Linear & dim=$10$, {bias} = True \\
         \bottomrule
    \end{tabular}
    \label{tab: mnist-addition-backbone}
\end{table}

In both \CBMs and DeepProbLog, each input digit $\vx^{(i)}$ is predicted independently and mapped to a $10$-dimensional bottleneck $\vz^{(i)}$.  Then, the two encodings are stacked together, obtaining the overall representation $\vz = (\vz^{(1)}, \vz^{(2)})$.

The classifier (top layer) of the \CBM is designed to predict the sum the $\vz^{(1)}$ and $\vz^{(2)}$.  A simple linear layer, which is the standard choice in \CBM \cite{koh2020concept}, is insufficient to successfully address the task. Therefore, we implemented the classifier via a bi-linear operation on the encodings, \ie:
$$ p_\theta(y| \vz^{(1)}, \vz^{(2)}) = \text{softmax}  \left( { \vz^{(1)}} \cdot W^y \vz^{(2)}\right)$$ 
where $W^y$ is a learnable class-specific $10 \times 10$ tensor of real entries, and the softmax is over all classes $y \in \{0,\dots,17\}$.

Conversely, the \textit{reasoning} (top) layer of \DeepProbLog is implemented as in \cref{eq:deepproblog-joint}.  Specifically, each $\vz^{(i)}$ encodes the logits of the probability $p_\theta (C_i | \vx^{(i)})$ for the $i$-th digit, which we convert into a categorical probability distribution through a softmax activation.

\underline{\CLEFVR:}  In a first step, we used Faster-RCNN~\citep{ren2015faster} to extract the bounding boxes associated to objects in all images.  The bounding box predictor is a pretrained MobileNet ~\citep{Howard2017MobileNetsEC} fine-tuned on training images from the first task only, using the ground-truth bounding boxes of \CLEFVR images.  The bounding box predictor is kept frozen in all successive tasks.  We discarded all examples $\vx_i$ with less than $2$ predicted bounding boxes.  Concept supervision was transferred from the ground-truth bounding boxes to the predicted ones based on overlap.

We scale each predicted bounding box to an image of size $28 \times 28 \times 3$ which is then passed to the encoder, implemented once again using a CNN with dropout with $p=50\%$ after each convolution layer and normal noise $\epsilon \sim {\cal N} (0, 0.1)$ added to the final output.
The architecture is the same as in \cref{tab: mnist-addition-backbone}, except that the input has depth $3$ instead of $1$ and that the bottleneck is $20$-dimensional, with $10$ dimensions allocated for the shape and $10$ for the color of the input object, each with its own softmax to produce shape and color probability distributions.
The \DeepProbLog \textit{reasoning} layer is as in \cref{eq:deepproblog-joint}.



During inference, the prediction returns invalid ($\perp$) whenever the number of predicted bounding boxes is less than $2$. We counted invalid predictions as wrong predictions in all metrics evaluated in the test set. 

\subsection{Hyper-parameter Selection}

All continual strategies have been trained with the same number of epochs and buffer dimension.  The actual values depend on the specific benchmark:
$25$ epochs per task and a buffer size of $1000$ examples for \MNISTSeq, $100$ epochs and $1000$ examples for \MNISTShortcut, and $50$ epochs each task and $250$ examples for \CLEFVR.
In all experiments, we employed the Adam optimizer \cite{KingmaB14@adam} combined with exponential decay ($\gamma=0.95$).  The initial learning rate is restored at the beginning of a new task. 

For each data set, we optimized the weight of the concept supervision $w_c$ based on the Class-IL $(\vY)$ performance of \er using grid-search on the validation set (union of all tasks).
Then, for each strategy, we selected the best learning rate and strategy-specific hyperparameters through a grid-search on the validation set, so as to optimize Class-IL $(\vY)$ on a single random seed.  The learning rate was chosen from the range of $[10^{-5}, 10^{-2}]$.  The exact values can be found in the source code.

\section{NeSy-CL Benchmarks} \label{sec:app-datasets}

In this section, we provide a more detailed description of the benchmarks introduced in \cref{sec:experiments}. 

\subsection{\MNISTSeq}

We derived \MNISTSeq from the \MNISTAdd data set of~\citet{manhaeve2018deepproblog}.  Here, the knowledge encodes the following constraint:
\[
    \textstyle
    \BK = \forall c_1,c_2 \in \{0,\dots,9\} \; (C_1 = c_1 \land C_2 = c_2) \implies Y = (c_1 + c_2)
\]
for a total of $19$ possible sums.  \MNISTSeq is both \textit{label-incremental} and \textit{concept-incremental}; in each task only two possible sums appear, obtaining in total $9$ tasks.  Specifically:
\begin{itemize}
    \item[] \textbf{Task $1$:} $Y \in \{0,1\}$ and $C \in \{\MZero, \MOne \}$;
    \item[] \textbf{Task $2$:} $Y \in \{2,3\}$ and $C \in \{\MZero, \MOne, \MTwo, \MThree \}$;
    \item[] \textbf{Task $3$:} $Y \in \{4,5\}$ and $C \in \{\MZero, \MOne, \MTwo, \MThree, \MFour, \MFive \}$;
    \item[] \textbf{Task $4$:} $Y \in \{6,7\}$ and $C \in \{\MZero, \MOne, \MTwo, \MThree, \MFour, \MFive, \MSix, \MSeven  \}$;
    \item[] \textbf{Task $5$:} $Y \in \{8, 9\}$ and $C \in \{\MZero, \MOne, \MTwo, \MThree, \MFour, \MFive, \MSix, \MSeven, \MEight, \MNine \}$;
    \item[] \textbf{Task $6$:} $Y \in \{10,11\}$ and $C \in \{\MOne, \MTwo, \MThree,  \MFour, \MFive, \MSix, \MSeven, \MEight, \MNine \}$;
    \item[] \textbf{Task $7$:} $Y \in \{12, 13\}$ and $C \in \{\MThree, \MFour, \MFive, \MSix, \MSeven, \MEight, \MNine \}$;
    \item[] \textbf{Task $8$:} $Y \in \{14,15 \}$ and $C \in \{ \MFive, \MSix, \MSeven, \MEight, \MNine \}$;
    \item[] \textbf{Task $9$:} $Y \in \{16,17 \}$ and $C \in \{\MSeven, \MEight, \MNine \}$;
\end{itemize}
In total, the data set counts $42$k training and $6$k test examples.

\subsection{\MNISTShortcut}

This benchmark is a case-study composed of two task, built considering the following constraints:
\begin{itemize}
    \item In the first task, we present only even digits and four possible sums:  (i) $\MZero + \MSix = 6$, (ii) $\MFour + \MSix = 10$, (iii) $\MTwo + \MEight = 10$, and (iv) $\MFour + \MEight = 12$. 
    \item In the second task, only odd numbers are considered, \ie $C \in \{ \MOne, \MThree, \MFive, \MSeven, \MNine \}$, and all their possible sums.
\end{itemize}
The rationale behind this construction is that it makes it possible to satisfy the knowledge in both tasks by leveraging reasoning shortcuts, and specifically those described in \cref{sec:shortcut-solutions}.

The overall data set contains $13.8$k training examples and $2$k test data.  We also collected an OOD test set containing examples not appearing in the training, validation and test sets, which comprise sums of odd and even digits, \eg $\MZero + \MOne = 1$, and unseen combinations of even numbers, \eg $\MEight+\MEight=16$.

\subsection{\CLEFVR} \label{sec:clevr_dataset}

Following \citet{stammer2021right}, the \CLEFVR data set was generated using Blender~\citep{blender}, using additional objects from \citep{li2022qlevr} as well as three custom shapes.

We generated different objects with $10$ possible shapes and colors, and with $2$ free variations on material $\{ {\tt rubber}, {\tt metal} \}$ and size $\{ {\tt small}, {\tt big} \}$, that play no role in determining the label. Each image is of size $256 \times 430  \times 3$, and contains two non-overlapping objects over a light-gray background.

\begin{figure}[!t]
    \centering
    \begin{tabular}{ccccc} 
     \sc task= 0 & \sc task = 1 &\sc task=2 &\sc task=3 &  \sc task=4  \\
     \includegraphics[width=0.15\textwidth]{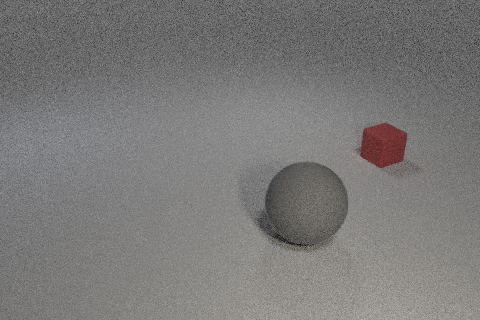} &
    \includegraphics[width=0.15\textwidth]{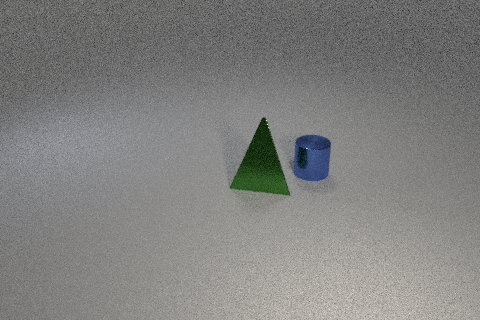} &
    \includegraphics[width=0.15\textwidth]{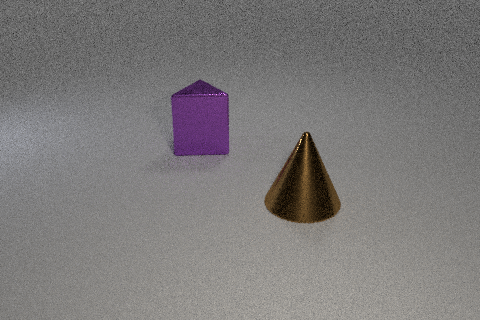} &
    \includegraphics[width=0.15\textwidth]{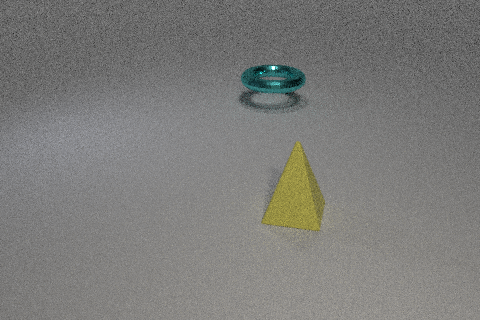} &
    \includegraphics[width=0.15\textwidth]{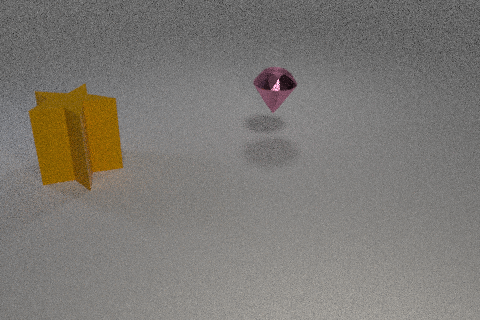} \\
    
    \includegraphics[width=0.15\textwidth]{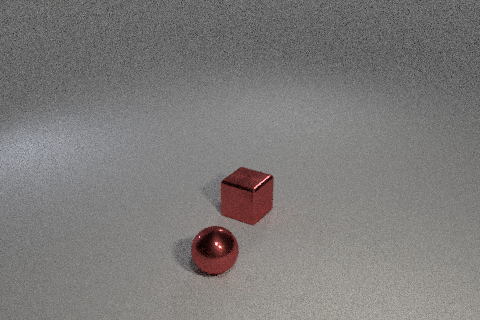} &
    \includegraphics[width=0.15\textwidth]{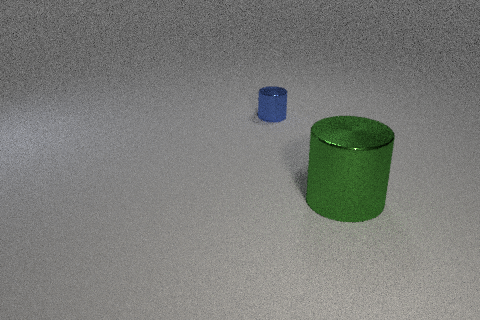} &
    \includegraphics[width=0.15\textwidth]{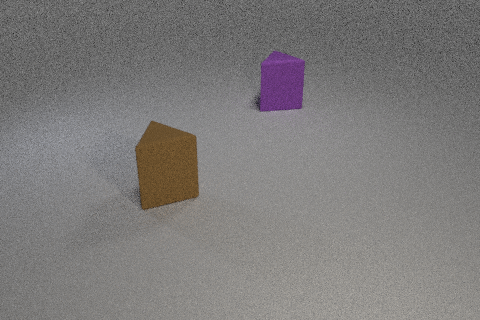} &
    \includegraphics[width=0.15\textwidth]{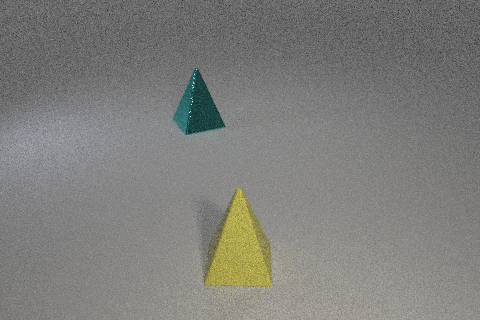} &
    \includegraphics[width=0.15\textwidth]{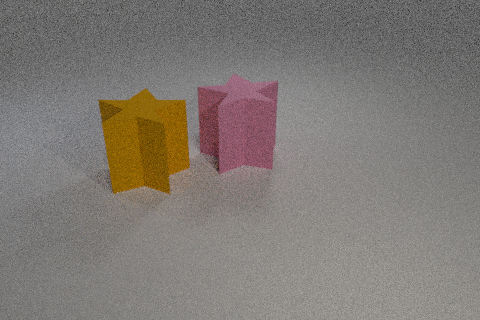} \\
    
    \includegraphics[width=0.15\textwidth]{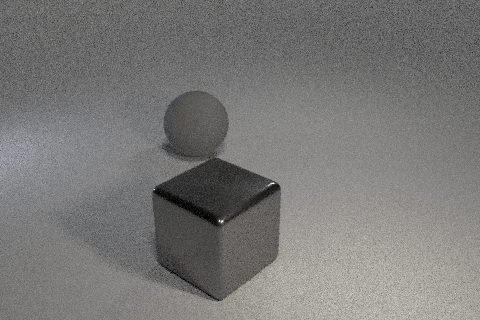} &
    \includegraphics[width=0.15\textwidth]{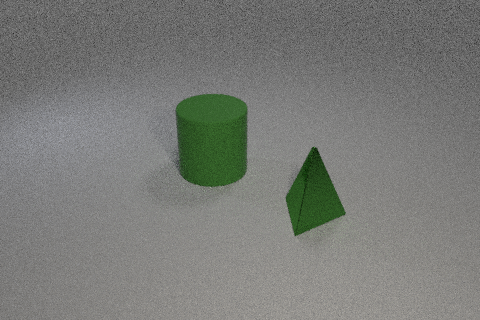} &
    \includegraphics[width=0.15\textwidth]{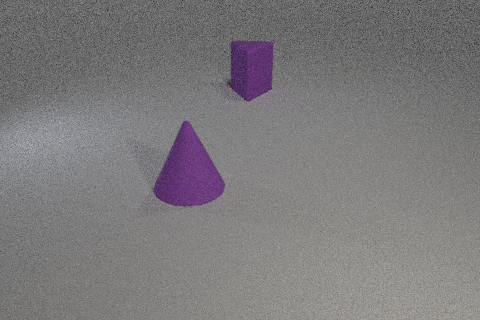} &
    \includegraphics[width=0.15\textwidth]{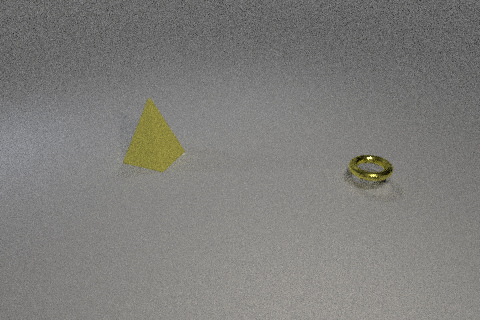} &
    \includegraphics[width=0.15\textwidth]{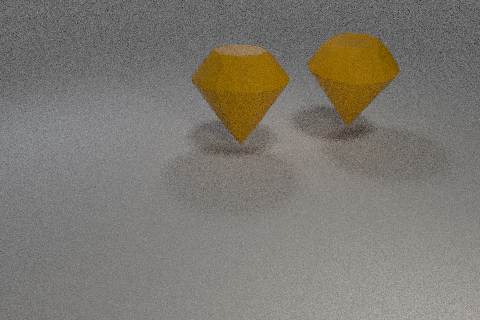} \\
    \end{tabular}
    \caption{Examples images for each tasks in the \CLEFVR benchmark.}
    \label{fig:examples_clevr}
\end{figure}

\cref{tab:clevr-task-design} reports what combinations of objects appear in each task.  All tasks are composed of $1.1$k training examples, $100$ validation examples, and $500$ test examples. Each task includes only two possible shapes (out of ten) and two colors (out of ten), without any overlap between tasks.  An illustration is given in \cref{fig:examples_clevr}.

The knowledge $\BK = \BK' \land \BK'' \land \BK'''$ encodes the following constraints:
\begin{align}
    \BK'
        & = (C_{{\tt shape},1} = C_{{\tt shape},2}) \iff {\tt same\_shape}
    \\
    \BK''
        & = (C_{{\tt color},1} = C_{{\tt color},2}) \iff {\tt same\_color}
    \\
    \BK'''
        & = ({\tt same\_shape} \land {\tt same\_color}) \iff {\tt same}
\end{align}
with three output variables $Y_1 = {\tt same\_shape}$, $Y_2 = {\tt same\_color}$ and $Y_3 = {\tt same}$, giving rise to four mutually exclusive classes: 0 = different shape and color, 1 = same shape and different color, 2 = different shape and same color, 3 = same shape and same color.

We also generated an additional OOD test set, comprising $300$ images depicting \textit{unseen combination} of training objects, \eg ${\tt red squares}$ and ${\tt pink diamonds}$ (which occur in none of the tasks).  All OOD examples all have label $\vY = (0, 0, 0)$.

All generated images come with ground-truth bounding boxes annotated with the properties (\ie concepts) of the objects they contain, as well as annotations for $Y_1$, $Y_2$, and $Y_3$.
The concept annotations are transferred to the bounding boxes predicted by Faster R-CNN during pre-processing, cf. \cref{sec:implementation-details}.

\begin{table}[!h]
    \centering
    \caption{Task Organization in \CLEFVR}
    \input{tables/clevr_setup}
    \label{tab:clevr-task-design}
\end{table}

\section{Shortcut Solutions in \MNISTShortcut and \CLEFVR}
\label{sec:shortcut-solutions}

In this section, we provide a more detailed account on the shortcut solutions for the continual scenarios introduced.

\subsection{Reasoning Shortcuts Due to Low-Level Correlations}

Before proceeding, we observe that simply predicting multiple concepts jointly by a single neural network is sufficient to enable reasoning shortcuts.  Intuitively, this happens because, since the network has access to all properties of all objects, it can automatically exploit correlations between them to satisfy the knowledge without the need for extracting any ``proper'' concepts.  For instance, in \MNISTAdd the knowledge can simply group pairs of digits (both of which it has access to) into two single combinations of concepts that always yield the right labels.
To see this, consider the following example:

\begin{example}
\textit{Consider a single \MNISTAdd task consisting of digits summing to either $2$ or $3$. By \cref{thm:ll-optima-satisfy-bk}, $\Theta^*(\BK, \dataset)$ contains $\varphi$ with $p_\varphi(\vC \mid \vx) \not\equiv p(\vC \mid \vx)$, such that:
\begin{align*}
    p_\varphi(C_1 \mid \vx)
        & = \Ind{C_1 = 2}
    \\
    p_\varphi(C_2 \mid \vx)
        & = \begin{cases}
                \Ind{C_2 = 0}
                    & \text{if $\vx = (\MZero, \MTwo) \ \text{or} \ (\MOne, \MOne)$}
                \\
                \Ind{C_2 = 1}
                    & \text{otherwise}
            \end{cases}
\end{align*}
This distribution fits the data perfectly and is consistent with the knowledge, and thus cannot be distinguished from the ground-truth distribution based on likelihood alone.}
\end{example}

Notice that the two learned concepts ignore the value of the individual digits, and rather depend on the correlation between them.  in \MNISTAdd, it is straightforward to avoid this situation \textit{by construction} by simply processing the two digits independently.  The same can be done when processing objects in \CLEFVR.  This partially motivates our choice of neural architecture, described in \cref{sec:implementation-details}.  More precisely, in \MNISTAdd the adopted architecture factorizes the joint probability distribution on the concepts in:
\begin{equation}
    p(\vc_1, \vc_2 | \vx^{(1)}, \vx^{(2)}) = p(\vc_1 | \vx^{(1)}) \cdot p(\vc_2 | \vx^{(2)})
\end{equation}

While this is sufficient to guarantee independence among distinct objects, the prior knowledge can still admit reasoning shortcuts. Next, we describe the concrete reasoning shortcuts appearing in \MNISTShortcut and \CLEFVR.

\subsection{Shortcuts in \MNISTShortcut}

\begin{figure}[!t]
    \centering
    \begin{tabular}{cc}
    \textsc{Offline} @ $0 \%$ & \textsc{Replay Methods} @ $10 \%$ \\
    \includegraphics[width=0.4\textwidth]{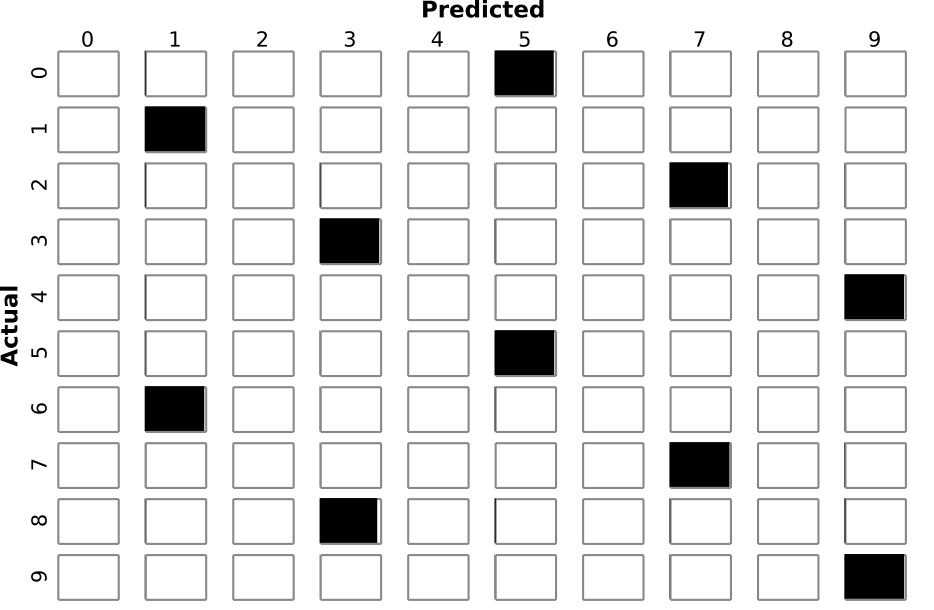}     &  
    \includegraphics[width=0.4\textwidth]{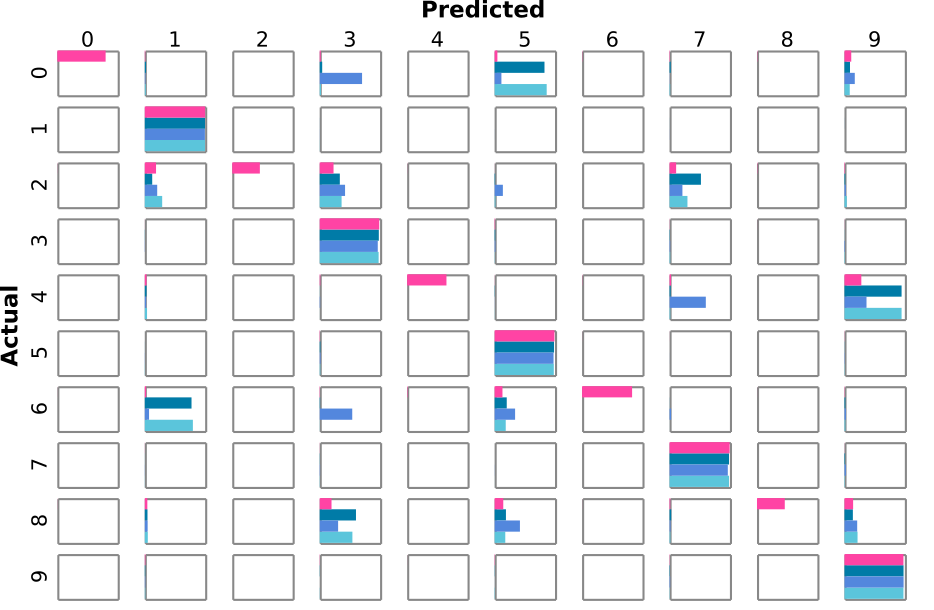}
    \end{tabular}
    \caption{Confusion matrices of the learned concepts at the end of the very last task, for a single run.
    \textbf{Left}: The confusion matrix on concepts for \offline without concept supervision. It shows that it is very likely to opt for a reasoning shortcut.
    \textbf{Right}: Confusion matrices on a single run for all replay-based methods. Here, \method is pictured in \textbf{\textcolor{WildStrawberry}{red}}, while \der, \derpp, and \er in shades of \textbf{\textcolor{Cerulean}{blue}}. Only \method retains the correct semantics of the concepts, whereas the other replay methods are very likely to pick the shortcut of \offline.} 
    \label{fig:confusion-matrices-shortcut}
\end{figure}

In order to understand reasoning shortcuts in \MNISTShortcut, it is useful to view the sums as constraints on the possible values attributed to each concept $C_j$.  From this perspective, reasoning shortcuts occur whenever the combinations of digits present in the training data (which are five in each task) are insufficient to uniquely determine the four possible sums.



In particular, the problem of assigning the intended semantics of each learned concept can be expressed as a system of $4$ linear equations with $5$ variables, which in task $1$ of \MNISTSeq can be written as:
\begin{equation}
\systeme{ c_0 + c_6 = 6, c_4 + c_6 = 10, c_2 + c_8 =10, c_4 + c_8 = 12 }
\end{equation}
The first equation states that whatever values are assigned to the concepts that fire when a \MZero and a \MSix are present in the input $\vx$, have to satisfy the condition that their sum is $6$ (in all examples in which they appear).
It should be clear that the linear system is undetermined and infinitely many real solutions can be found for $c_i$, all of which except one do not capture the intended semantics of digits.

One of these unintended solutions is very often picked by label-replay strategies.  Specifically, the input $\MFour$ can be easily confused for a $9$, due to input similarity, which brings the model towards the assignment $c_4=9$.  This immediately yields the following unintentional mappings for all other digits: $c_0=5$, $c_2=7$, $c_6=1$, and $c_8=3$. 
This reasoning shortcut is often found when training offline on this task and also when training on both tasks of \MNISTShortcut, as done in our experiments, by \er, \der, and \derpp, as shown in \cref{fig:confusion-matrices-shortcut}. 


\subsection{Shortcuts in \CLEFVR}

Several shortcut exist in \CLEFVR.  Recall that:
\begin{itemize}

    \item \CLEFVR is composed of 5 tasks, with four possible outcomes (\textit{different objects}, \textit{same shapes}, \textit{same colors}, and \textit{same objects}).

    \item In each task only two possible shapes and colors are observed, and are never seen again in other tasks.

\end{itemize}
In order to correctly classify the ${\tt same\_color}$ and ${\tt same\_shape}$ labels, the model must acquire at least two distinct concepts for shape and two distinct concepts for colors in each task, but the knowledge provides little guidance as to \textit{which} shapes or colors need to be associated to the input objects.  This leaves ample room for reasoning shortcuts.

Let us focus on shapes only.  Letting $\cal S$ be the set of the $10$ possible shapes $s_i$, the model needs at least $10 \cdot (10-1)/2$ different negative examples of the knowledge to uniquely identify  all possible shapes (up to permutation), one for each pair of mismatching shapes.
Consider the map $\pi: \calS \to \calS$ mapping from ground-truth shape of the input object to the learned concept for shape.  Ideally, we would like $\pi$ to be injective, so that no two distinct ground-truth shapes are mapped to the same learned shape.


Consider a task with $4$ possible shapes $s_1 = {\tt cube}$, $s_2 = {\tt cone}$, $s_3 = {\tt cylinder}$, and $s_4 = {\tt toroid}$.  In order to guarantee injectivity, the data has to include enough combinations of shapes so that the map $\pi$ satisfies the following condition:
\begin{equation} \label{eq:condition-for-clevr}
\begin{aligned}
    & \pi(s_1) \neq \pi(s_2), \quad \pi(s_1) \neq \pi(s_3), \quad \pi(s_1) \neq \pi(s_4), \\
    & \pi(s_2) \neq \pi(s_3), \quad \pi(s_2) \neq \pi(s_4), \\
    & \pi(s_3) \neq \pi(s_4) 
\end{aligned}
\end{equation}
Notice that this map \textit{is} injective, in the sense that $s_i \neq s_j \implies \pi(s_i) \neq \pi(s_j), \; \forall i\neq j$.
All tasks in \CLEFVR, however, are designed to distinguishing between only \textit{two} possible shapes (and colors), hence the condition in \cref{eq:condition-for-clevr} is never satisfied.  This is what allows for reasoning shortcuts.

\begin{figure}[!t]
    \centering
    \begin{tabular}{cccc}
        & \sc Shapes & \sc Colors & \sc Labels  \\ \cmidrule{2-4} 
        \rotatebox{90}{\hspace{2em}  \scriptsize \sc Ground-Truth} &
        \includegraphics[width=0.25 \textwidth]{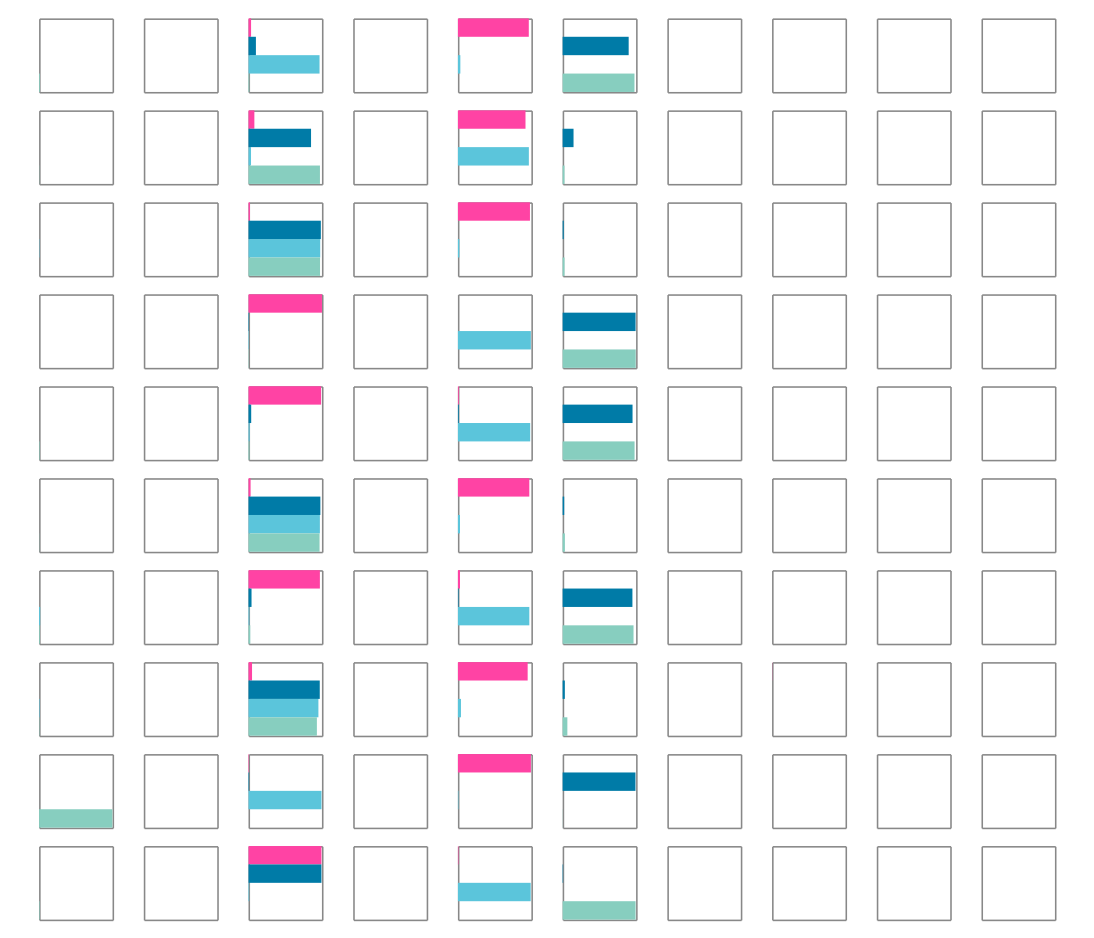} & 
        \includegraphics[width=0.25 \textwidth]{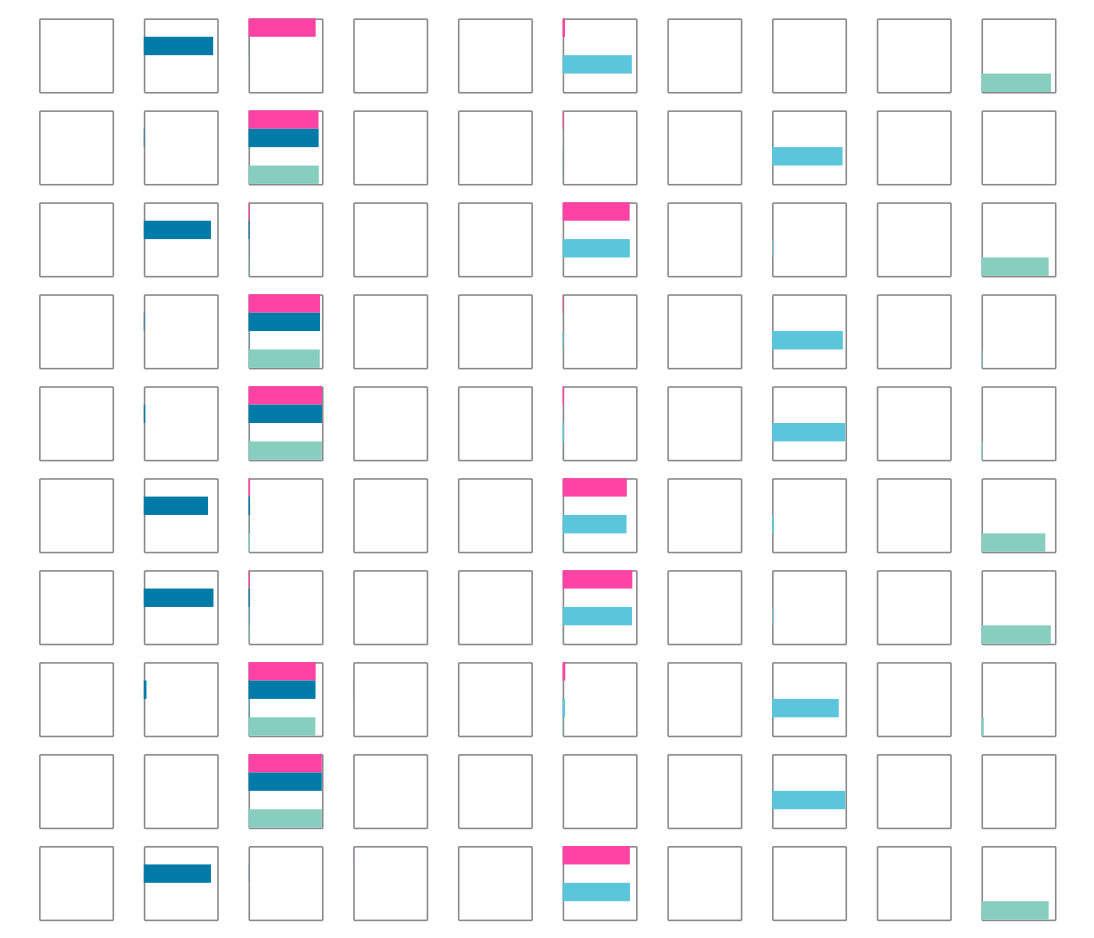} &
        \includegraphics[width=0.25 \textwidth]{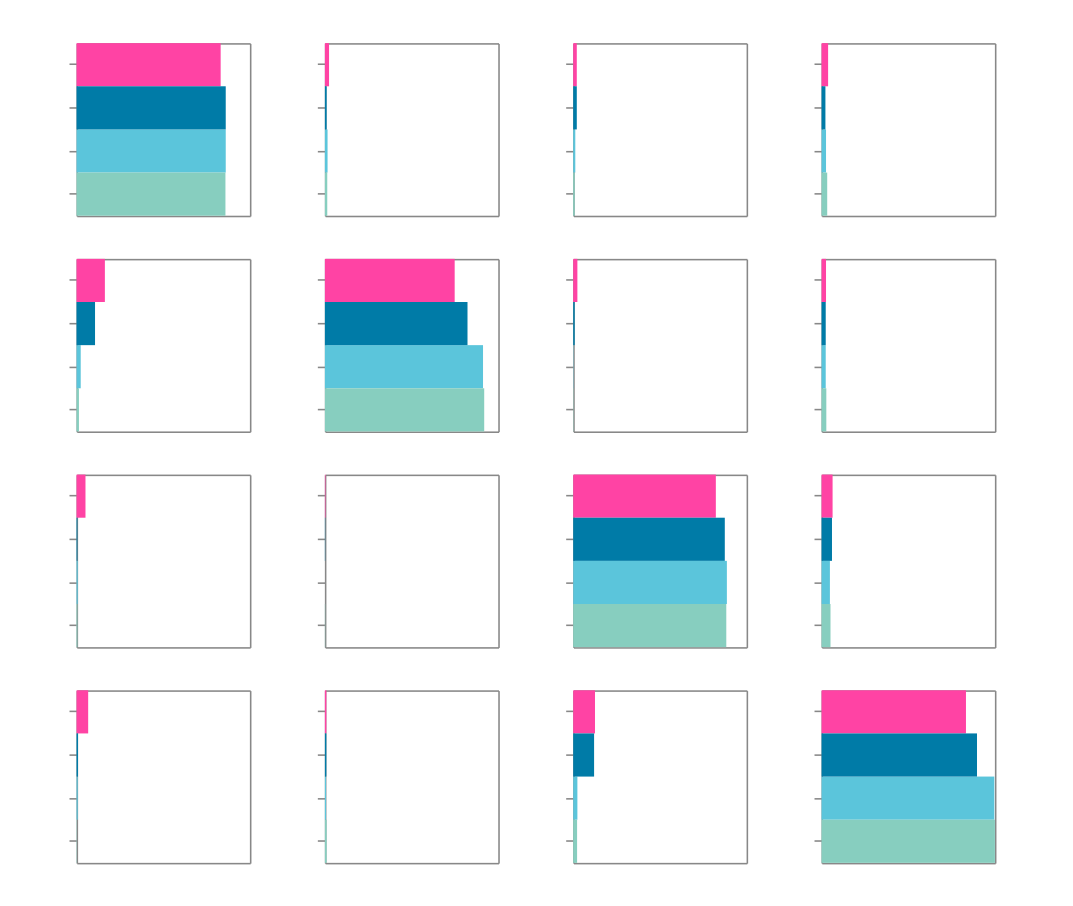} \\
        & \scriptsize \textsc{Predicted} & \scriptsize  \textsc{Predicted} &  \scriptsize \textsc{Predicted}    
    \end{tabular}
    \caption{Confusion matrices for \method (in \textbf{\textcolor{WildStrawberry}{red}}) and the other replay strategies (in shades of \textbf{\textcolor{Cerulean}{blue}}) in \CLEFVR with no concept supervision, at the end of all tasks. All methods fails to attribute the correct semantics to the concepts and learn, instead, a shortcut which optimizes the Class-IL ($\vY$).}
    \label{fig:confusions-no-sup-clevr}
\end{figure}

As a matter of fact, without concept supervision, all values for shapes and colors are equally likely to be picked up to solve the task. We observed this phenomenon in the case of $0\%$ supervision reported in \cref{fig:confusions-no-sup-clevr}.


\newpage
\section{Additional Results and Metrics} 
\label{sec:additional-results}

\subsection{Time Overhead and Memory Occupation of \method}

\changed{We evaluate the time overhead and the memory requirements of \method compared to other replay-based strategies in \cref{tab:memory-time-replay}.
All these strategies impose an additional time overhead due to the selection and replay of past examples.} 

\changed{We measured the time required for completing a single epoch in the first task of \MNISTAdd and \CLEFVR. 
\naive provides the lower bound for the computation time, as it just fine-tunes the model on the new examples. \er in \MNISTAdd strategy requires almost $\sim 0.1 \,$s more than \naive for storing and replaying. In \CLEFVR, however, the gap is less evident between the two strategies.
For \MNISTSeq, the time per epoch of \method amounts to $\sim 0.476 \,$s, which is comparable to that of \der $\sim 0.475 \,$s and lower than that of \derpp $\sim 0.667 \,$ s. For \CLEFVR \method requires $0.439 \,$s per epoch, slightly increasing compared to \der $\sim 0.412 \,$s, but still being lower w.r.t. \derpp $\sim 0.482 \,$s.}

\changed{In terms of memory occupation, \method requires slightly more memory than other replay strategies, due to the need of storing the \textit{logits} of the learned concepts. However, most of the overhead is due to storing the instances $\vx$ themselves, which is the very same for all strategies and amounts to $6.272$ Mb for \MNISTAdd and $1.568$ Mb for \CLEFVR.}

\begin{table}[!t]
    \centering
    \scriptsize
    \include{tables/memory-time}
    \caption{\changed{\textbf{Time per epoch and memory occupation} for \MNISTAdd and \CLEFVR.}}
    \label{tab:memory-time-replay}
\end{table}

\subsection{\MNISTSeq}

We report in \cref{tab:full-mnist-seq} results for all competing strategies and performance measures used (including those omitted in the main text), namely
 Class-IL and Task-IL on labels and concepts (denoted as $\vY$ and $\vC$, respectively), backward transfer (BWT), and forward transfer (FWT). 

\begin{table}[!h]
    \begin{scriptsize}
        \begin{center}
            \input{tables/new-CBM-vs-NeSy-fixed}
        \end{center}
    \end{scriptsize}
     \caption{\textbf{Additional results for \MNISTSeq.}
     }
    \label{tab:full-mnist-seq}
\end{table}

These results confirm the ones reported in \cref{sec:experiments}.  Specifically, \method attains higher performance w.r.t. all metrics in both \CBM and \DeepProbLog.  They also show that \method outperforms all other approaches in terms of BWT when paired with DeepProbLog, and is the runner-up with \CBMs.

The method with the best BWT on \CBM is \lwf, which however displays a pathological learning behavior, as made clear by the fact that it is the only method to have \textit{negative} forward transfer.  The reason is that \lwf struggles to learn the first few tasks properly, but performs reasonably on the latter ones.  

Surprisingly, this oddball behavior is sufficient for \lwf to beat the baselines (\naive and \restart) in terms of Class-IL $\vY$ (at around $18\%$), but not enough to outperform \method, and also yields the aforementioned issue with FWT.  We stress that this implies that \lwf generalizes to forward tasks \textit{worse than random}. 


\subsection{\MNISTShortcut}

We report here in \cref{tab:mnist-shortcut-full-results}, all results obtained for all strategies in \MNISTShortcut. We performed the comparison adopting only DeepProbLog with increasing amount of concept supervision.

\begin{table*}[!h]
    \begin{scriptsize}
        \begin{center}
        \input{tables/SHORTCUT-ALL-RESULTS}

        \end{center}
    \end{scriptsize}
    \caption{\MNISTShortcut Additional results.}
    \label{tab:mnist-shortcut-full-results}
\end{table*}

With $0 \%$ concept supervision, all methods perform poorly, \ie worse or comparably to \naive and \restart in terms of Class-IL.
Variance is also quite large for \ewc, \er, \der, \derpp, and \method.
The sole exception is \lwf, which obtains higher performance and small variance in label classification.
On the other hand, the results in OOD accuracy are all below $13 \%$, indicating that no strategy can successfully identify high-quality concepts that transfer across tasks.

The picture at $1\%$ and $10\%$ supervision is very similar: \method outperforms all approaches in terms of concept quality and OOD accuracy by a large margin, while the other approaches pick up the reasoning shortcut, thus achieving high label accuracy only.

\subsubsection{\CLEFVR}

\begin{table*}[!h]
    \begin{scriptsize}
        \begin{center}
            \input{tables/CLEVR-ALL-RUNS}
        \end{center}
    \end{scriptsize}
    \caption{\CLEFVR Full results}
    \label{fig:complete-results-clevr}
\end{table*}

\begin{figure}[!t]
      \setlength{\tabcolsep}{3pt}
    \centering
    \begin{tabular}{cccccc}
       & \texttt{task}$=1$ & \texttt{task}$=2$ & \texttt{task}$=3$ & \texttt{task}$=4$ & \texttt{task}$=5$   \\
       \rotatebox{90}{\hspace{1.75em} \texttt{shapes}} &
        \includegraphics[width=0.17\textwidth]{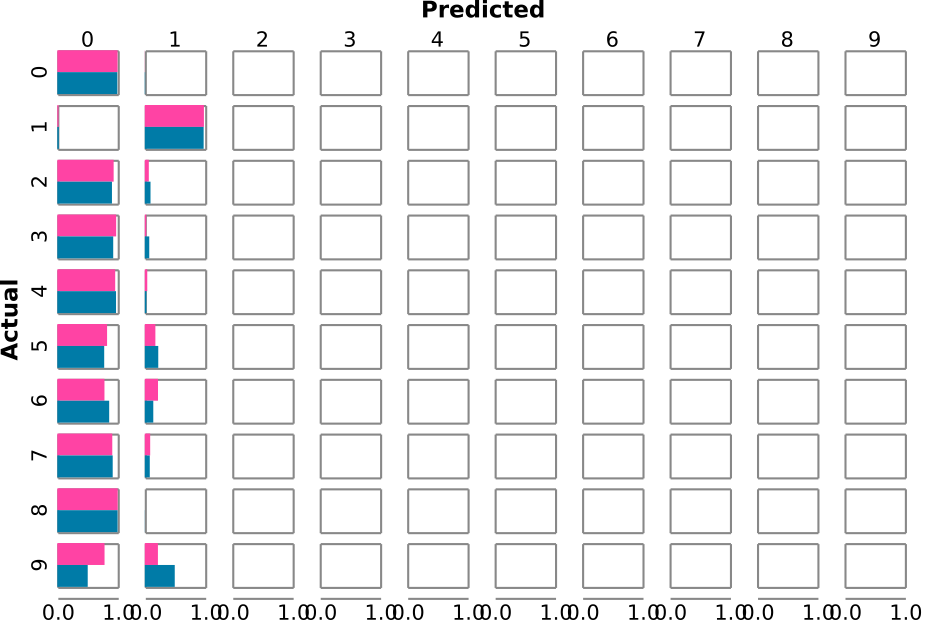} &
        \includegraphics[width=0.17\textwidth]{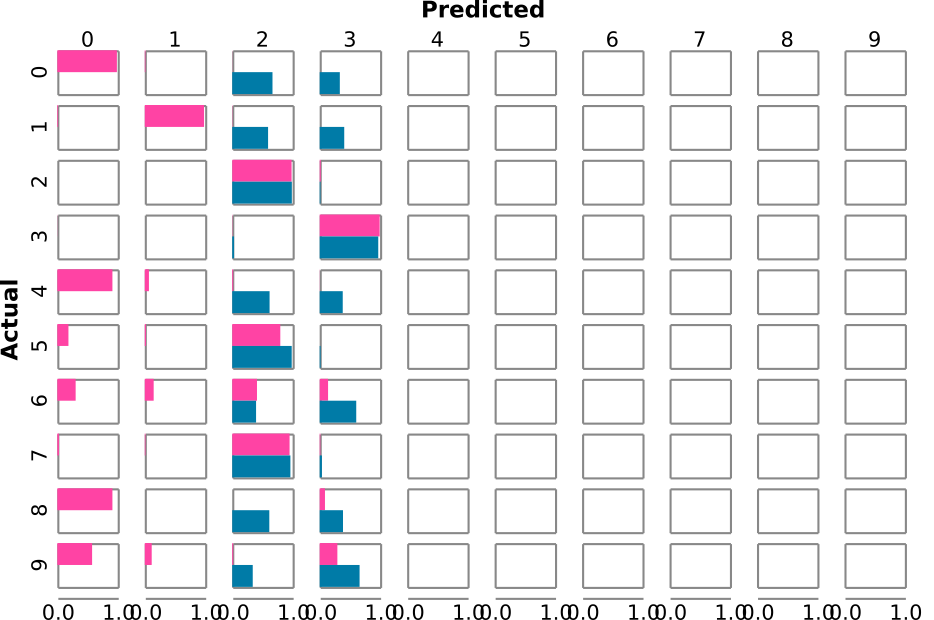} &
        \includegraphics[width=0.17\textwidth]{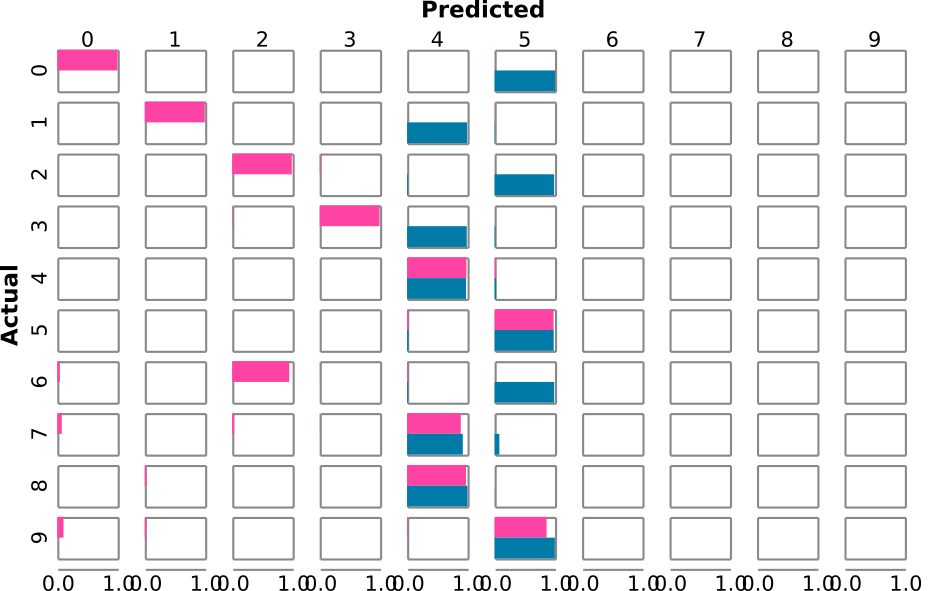} &
        \includegraphics[width=0.17\textwidth]{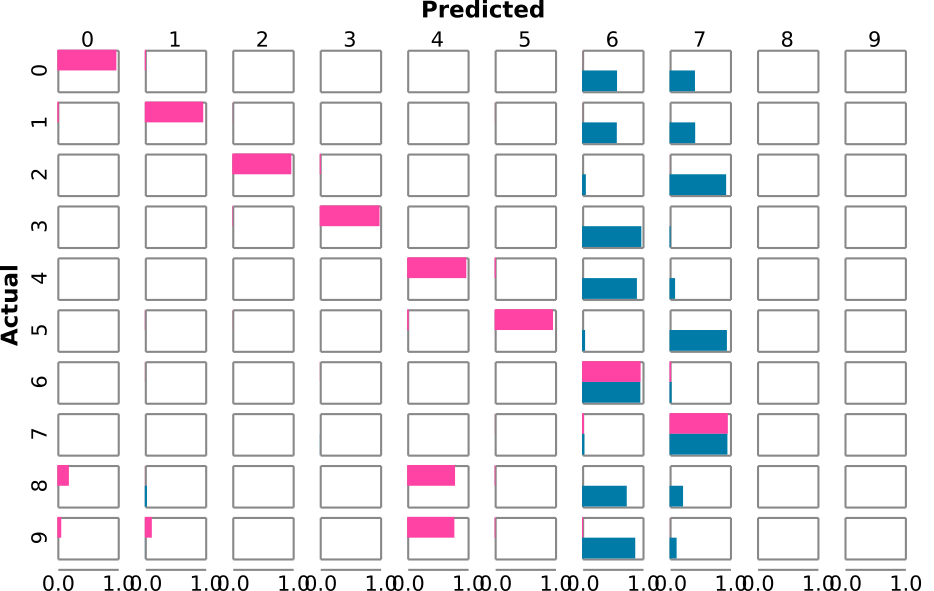} &
        \includegraphics[width=0.17\textwidth]{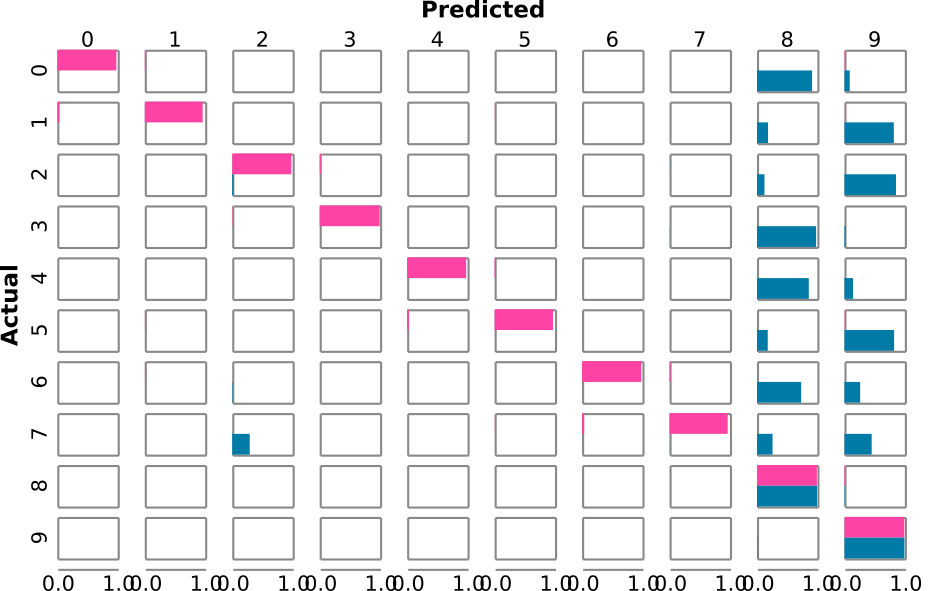}  \\
        \rotatebox{90}{\hspace{1.7em} \texttt{colors}} &
        \includegraphics[width=0.17\textwidth]{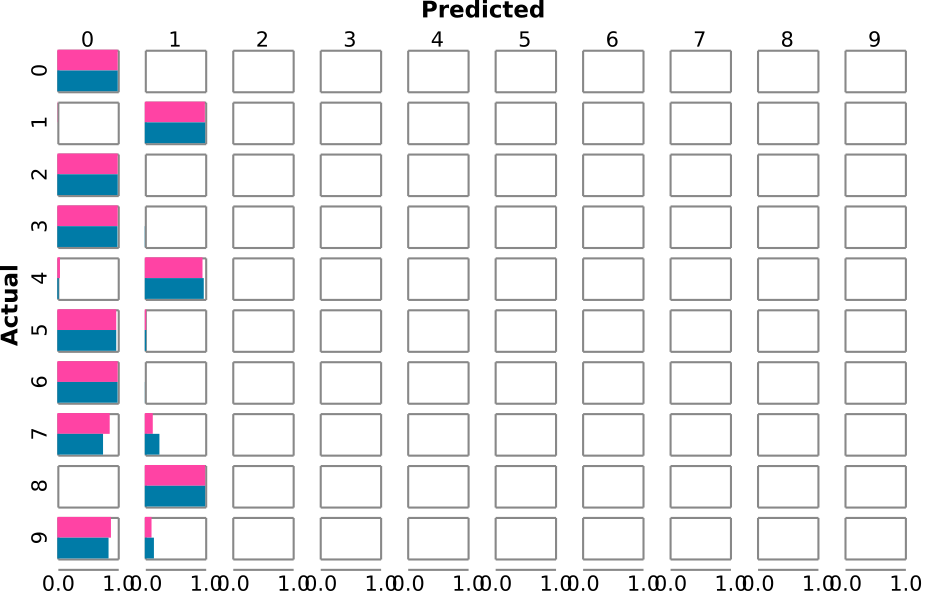} &
        \includegraphics[width=0.17\textwidth]{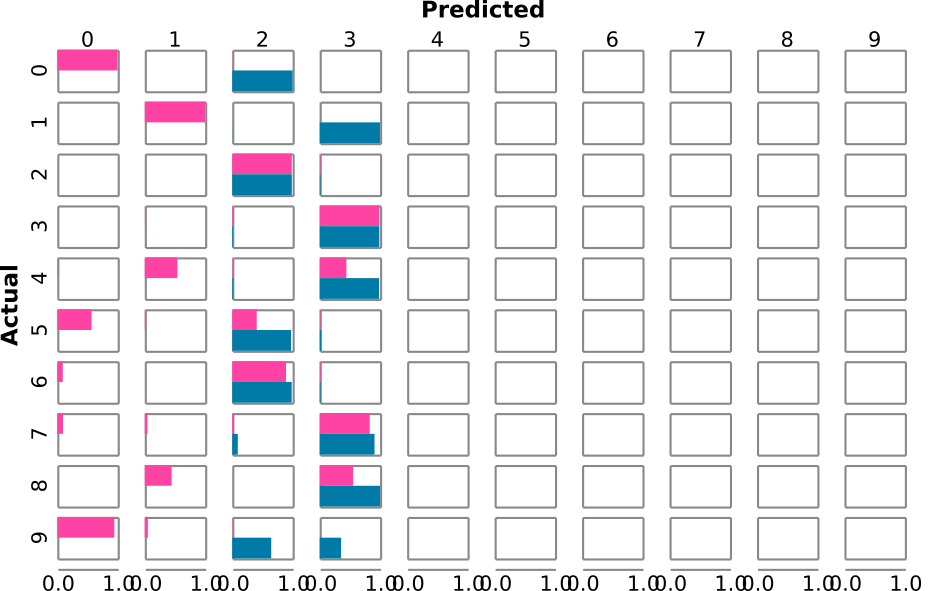} &
        \includegraphics[width=0.17\textwidth]{figures/confusions/clevr-confusion-color-3.png} &
        \includegraphics[width=0.17\textwidth]{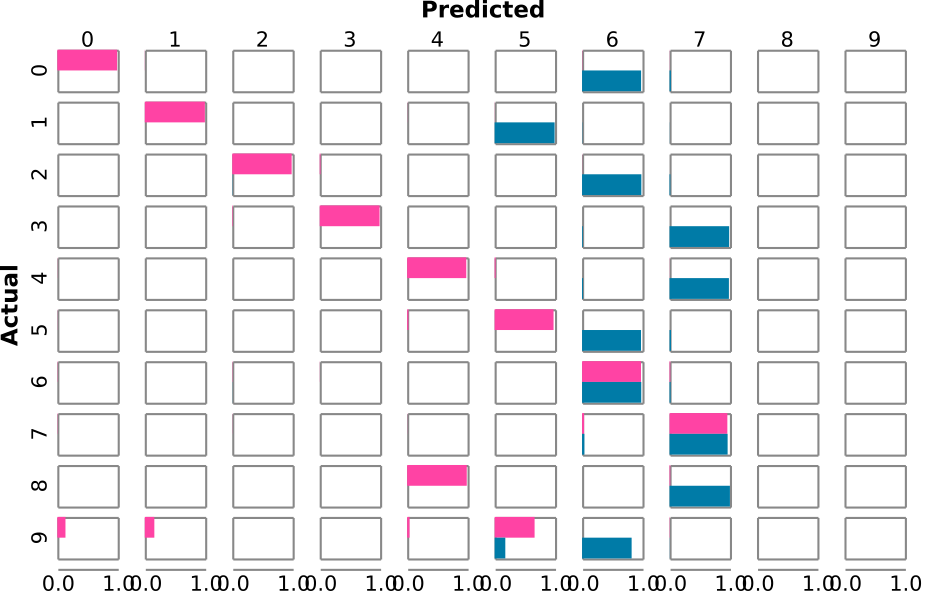} &
        \includegraphics[width=0.17\textwidth]{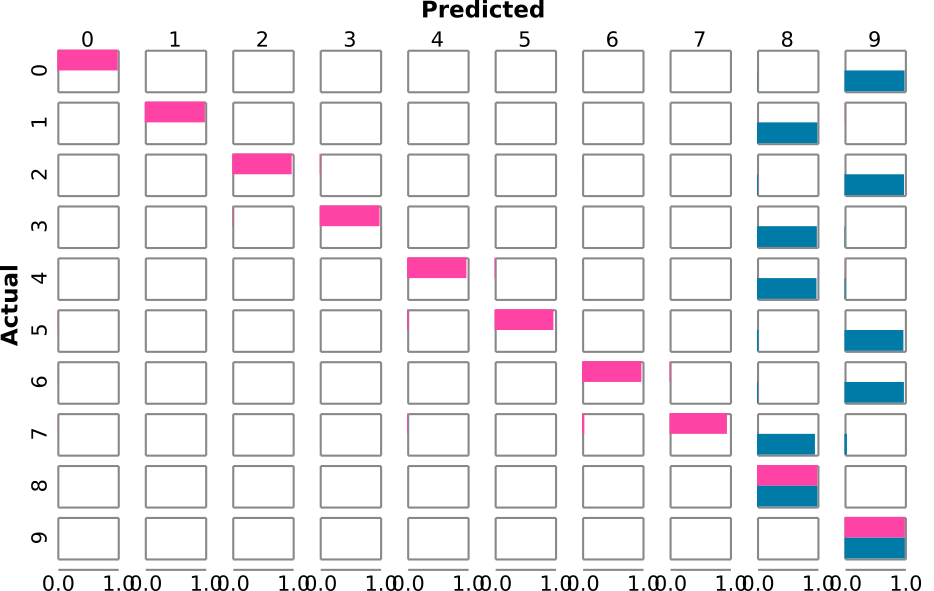} \\
    \end{tabular}
    \caption{Dynamics of confusion matrices for shapes and colors encodings on \CLEFVR with 10\% concept supervision.  \method in \textbf{\textcolor{WildStrawberry}{red}} preserves the correct concepts, while \der (in \textbf{\textcolor{Cerulean}{blue}}) suffers for shortcuts. Shape and color range in $\{0, \ldots, 9\}$.}
    \label{tab:confusion-matrices-clevr}
\end{figure}

The complete results for \CLEFVR, reported in \cref{fig:complete-results-clevr}, show the same trend as those on \MNISTShortcut.
For completeness, we report the evolution across tasks of concept confusion matrices for all methods in \cref{tab:confusion-matrices-clevr}.  The impact of the reasoning shortcut on \der, and its lack of impact on \method, are clearly visible.

\end{document}

%% file: tables/MNIST-ADD-CBM-vs-NESY.tex
\begin{tabular}{llccc}
    \toprule
                                                         & {\sc Strategy}  & {\sc Class-IL} $\vY$ ($\uparrow$) & {\sc Class-IL} $\vC$ ($\uparrow$)        & FWT ($\uparrow$) \\ \cmidrule{2-5}
\multirow{8}{*}{\rotatebox{90}{\sc CBM @ $10\%$}}        & \naive         & $11.71 \pm 0.8$                & $36.2 \pm 2.6$              & $7.5 \pm 0.3$ \\ 
                                                         & \restart       & $10.78 \pm 0.1$                & $29.7 \pm 0.1$              & $7.3 \pm 0.2$ \\
                                                         & \lwf         & $18.08 \pm 1.8$                & $63.2 \pm 4.4$              & $-4.7 \pm 1.1$ \\
                                                         & \ewc          & $11.57 \pm 0.6$                & $37.4 \pm 0.6$              & $7.6 \pm 0.4$ \\ 
                                                         & \er            & $13.29 \pm 0.4$                & $43.5 \pm 2.0$              & $13.4 \pm 1.6$ \\
                                                         & \der           & $18.63 \pm 2.5$                & $53.1 \pm 1.7$              & $15.7 \pm 0.9$ \\
                                                         & \derpp         & $18.17 \pm 1.6$                & $54.1 \pm 3.0$              & $16.6 \pm 1.8$ \\ 
                                                         & \method     & $\bm{38.0 \pm 1.9}$            & $\bm{78.1 \pm 2.5}$         & $\bm{29.0 \pm 4.8}$ \\ \cmidrule{2-5}
\multirow{8}{*}{\rotatebox{90}{\sc DeepProbLog }} & \naive         & $ 6.9 \pm 0.2$                 & $ 6.7 \pm 0.4$              & $ 6.2 \pm 0.2$ \\ 
                                                         & \restart      & $ 9.6 \pm 0.3$                 & $ 0.2 \pm 0.1$              & $ 6.9 \pm 0.8$ \\
                                                         & \lwf          & $ 6.8 \pm 0.5$                 & $10.8 \pm 4.6$              & $18.3 \pm 0.2$ \\ 
                                                         & \ewc          & $ 6.8 \pm 0.4$                 & $ 7.8 \pm 0.6$              & $ 6.1 \pm 0.3$ \\ 
                                                         & \er            & $44.3 \pm 9.7$                 & $62.0 \pm 8.6$              & $ 8.2 \pm 4.1$ \\ 
                                                         & \der          & ${68.3 \pm 9.4}$            & ${81.3 \pm 6.9}$               & $44.5 \pm 23.7$ \\ 
                                                         & \derpp         & $62.2 \pm 5.4$                 & $77.1 \pm 4.2$              & $27.1 \pm 5.2$ \\ 
                                                         & \method        & $\bm{71.9 \pm 2.9}$            & $\bm{84.5 \pm 1.9}$         & $\bm{83.2 \pm 0.9}$ \\ 
    \bottomrule
\end{tabular}

%% file: tables/clevr_setup.tex
\begin{tabular}{ccccc}
\toprule
{\sc Task}      &  {\sc Colors}                  & {\sc Shapes} \\
\midrule
$1$             & ${\tt red}$, ${\tt gray}$      & ${\tt sphere}$, ${\tt cube}$ \\
$2$             & ${\tt green}$, ${\tt blue}$    & ${\tt cylinder}$, ${\tt tetrahedron}$ \\
$3$             & ${\tt brown}$, ${\tt purple}$  & ${\tt cone}$, ${\tt triangular}$ ${\tt prism}$ \\
$4$             & ${\tt yellow}$, ${\tt cyan}$   & ${\tt pyramid}$, ${\tt toroid}$ \\
$5$             & ${\tt orange}$, ${\tt pink}$   & ${\tt diamond}$, ${\tt star}$ ${\tt prism}$ \\
\bottomrule
\end{tabular}

%% file: tables/memory-time.tex
\begin{tabular}{lccccc}
    \toprule
       & \multicolumn{2}{c}{\footnotesize \MNISTSeq}               &        & \multicolumn{2}{c}{\footnotesize \CLEFVR }                            \\
       & \multicolumn{2}{c}{\sc  (Buffer size = 1000)}         &        & \multicolumn{2}{c}{\sc  (Buffer size = 250)}                  \\
       \cmidrule{2-3} \cmidrule{5-6}
       & \sc Time per epoch & \textsc{Memory occupation}    &        & \sc Time per epoch & \textsc{Memory occupation} \\
\naive  & 0.238 s        & -                                &       & 0.335 s        & -                                     \\
\er     & 0.385 s        & 6.280 Mb                         &       & 0.344 s        & 1.570 Mb                              \\
\der    & 0.475 s        & 6.348 Mb                         &       & 0.412 s        & 1.587 Mb                              \\
\derpp  & 0.667 s        & 6.356 Mb                         &       & 0.482 s        & 1.589 Mb                              \\
\method & 0.476 s        & 6.360 Mb                         &       & 0.439 s        & 1.590 Mb   \\
\bottomrule
\end{tabular}

%% file: tables/new-CBM-vs-NeSy-fixed.tex
\begin{tabular}{llcclcclcc}
    \toprule
         &                & \multicolumn{2}{c}{{\sc Labels} ($\vY$)}       &                  & \multicolumn{2}{c}{{\sc Concepts} ($\vC$)}   & & \\
         \cmidrule{3-4} \cmidrule{6-7} \cmidrule{9-10}
                                                         & {\sc Strategy}  & {\sc Class-IL} ($\uparrow$)         & {\sc Task-IL}  ($\uparrow$)        & & {\sc Class-IL} ($\uparrow$)        & {\sc Task-IL} ($\uparrow$)  & & {\sc BTW} ($\uparrow$)      & {\sc FWT} ($\uparrow$) \\ \cmidrule{2-10}
\multirow{8}{*}{\rotatebox{90}{\sc CBM @ $10\%$}}        
                                                         & \naive         & $11.71 \pm 0.8$              & $27.5 \pm  3.9$             & & $36.2 \pm 2.6$              & $50.8 \pm 1.7$       & & $-91.1 \pm 1.2$       & $7.5 \pm 0.3$ \\ 
                                                         & \restart       & $10.78 \pm 0.1$              & $32.9 \pm  2.1$             & & $29.7 \pm 0.1$              & $43.1 \pm 0.6$       & & $-98.2 \pm 0.1$       & $7.3 \pm 0.2$ \\
                                                         & \lwf           & $18.08 \pm 1.8$              & $55.7 \pm  5.3$             & & $63.2 \pm 4.4$              & $78.6 \pm 2.5$       & & $\bm{-18.5 \pm 1.8}$  & $-4.7 \pm 1.1$ \\
                                                         & \ewc           & $11.57 \pm 0.6$              & $38.3 \pm 11.2$             & & $37.4 \pm 0.6$              & $52.1 \pm 1.7$       & & $-90.4 \pm 0.8$       & $7.6 \pm 0.4$ \\ 
                                                         & \er            & $13.29 \pm 0.4$              & $32.5 \pm  3.2$             & & $43.5 \pm 2.0$              & $67.7 \pm 3.3$       & & $-88.0 \pm 0.7$       & $13.4 \pm 1.6$ \\
                                                         & \der           & $18.63 \pm 2.5$              & $50.3 \pm  3.4$             & & $53.1 \pm 1.7$              & $73.1 \pm 2.0$       & & $-88.6 \pm 2.9$       & $15.7 \pm 0.9$ \\
                                                         & \derpp         & $18.17 \pm 1.6$              & $48.9 \pm  2.4$             & & $54.1 \pm 3.0$              & $73.1 \pm 4.5$       & & $-89.2 \pm 1.8$       & $16.6 \pm 1.8$ \\ 
                                                         & \method        & $\bm{38.0 \pm 1.9}$          & $\bm{78.4 \pm 3.8}$         & & $\bm{78.1 \pm 2.5}$     & $\bm{90.4 \pm 1.6}$      & & $-68.3 \pm 2.1$       & $\bm{29.0 \pm 4.8}$ \\ \cmidrule{2-10}
\multirow{8}{*}{\rotatebox{90}{\sc DeepProbLog @ $0\%$}} 
                                                         & \naive         & $ 6.9 \pm 0.2$               & $ 7.6 \pm 0.2$              & & $ 6.7 \pm 0.4$              & $16.4 \pm 1.0$       & & $-63.1 \pm 0.6$       & $ 6.2 \pm 0.2$ \\ 
                                                         & \restart       & $ 9.6 \pm 0.3$               & $22.2 \pm 0.8$              & & $ 0.2 \pm 0.1$              & $11.6 \pm 0.5$       & & $-69.5 \pm 1.9$       & $ 6.9 \pm 0.8$ \\
                                                         & \lwf           & $ 6.8 \pm 0.5$               & $11.0 \pm 1.6$              & & $10.8 \pm 4.6$              & $21.7 \pm 8.3$       & & $-71.2 \pm 5.2$       & $18.3 \pm 0.2$ \\ 
                                                         & \ewc           & $ 6.8 \pm 0.4$               & $ 7.8 \pm 0.4$             & & $ 7.8 \pm 0.6$              & $18.3 \pm 0.6$       & & $-62.9 \pm 0.4$       & $ 6.1 \pm 0.3$ \\ 
                                                         & \er            & $44.3 \pm 9.7$               & $81.2 \pm 6.5$             & & $62.0 \pm 8.6$              & $82.8 \pm 5.1$       & & $-48.3 \pm 8.8$       & $ 8.2 \pm 4.1$ \\ 
                                                         & \der           & ${68.3 \pm 9.4}$             & ${94.8 \pm 3.2}$           & & ${81.3 \pm 6.9}$            & ${93.8 \pm 3.0}$  & & ${-30.5 \pm 8.5}$  & $44.5 \pm 23.7$ \\ 
                                                         & \derpp         & $62.2 \pm 5.4$               & $93.5 \pm 2.1$              & & $77.1 \pm 4.2$              & $92.8 \pm 2.9$       & & $-34.8 \pm 5.6$       & $27.1 \pm 5.2$ \\ 
                                                         & \method        & $\bm{71.9 \pm 2.9}$          & $\bm{96.6 \pm 0.8}$         & & $\bm{84.5 \pm 1.9}$         & $\bm{95.4 \pm 0.4}$  & & $\bm{-28.7 \pm 3.2}$  & $\bm{83.2 \pm 0.9}$ \\ 
    \bottomrule
\end{tabular}

%% file: tables/SHORTCUT-ALL-RESULTS.tex
      \setlength{\tabcolsep}{3pt}
    \begin{tabular}{lccccccccccc}

            \toprule 
            & \multicolumn{3}{c}{\sc Supervision $0 \%$ } & & \multicolumn{3}{c}{\sc Supervision $1 \%$ } & & \multicolumn{3}{c}{\sc Supervision $10 \%$ } \\ 
            \cmidrule{2-4} \cmidrule{6-8} \cmidrule{10-12}
             & \sc Class-IL($\vY$) & \sc Class-IL($\vC$) & \sc {OOD  ($\vY$)} & ~ & \sc Class-IL($\vY$) & \sc Class-IL($\vC$) & \sc {OOD  ($\vY$)} & ~ & \sc Class-IL($\vY$) & \sc Class-IL($\vC$) & \sc {OOD  ($\vY$)} \\ 
            \naive & 59.9 $\pm$ 3.2 & 49.4 $\pm$ 0.4 & 10.7 $\pm$ 2.3 & ~ & 57.5 $\pm$ 0.5 & 49.2 $\pm$ 0.1 & 12.9 $\pm$ 0.4 & ~ & 59.9 $\pm$ 0.9 & 49.4 $\pm$ 0.1 & 12.3 $\pm$ 0.5 \\ 
            \restart & 59.1 $\pm$ 0.6 & \bf 49.7 $\pm$ 0.1 & \bf 12.9 $\pm$ 0.4 & ~ & 58.1 $\pm$ 1.6 & 49.3 $\pm$ 0.1 & 12.9 $\pm$ 0.5 & ~ & 59.1 $\pm$ 0.9 & 49.6 $\pm$ 0.1 & 13.2 $\pm$ 0.6 \\ 
            \ewc & 55.9 $\pm$ 8.9 & 45.6 $\pm$ 11.4 & 12.5 $\pm$ 1.4 & ~ & 59.2 $\pm$ 1.5 & 49.5 $\pm$ 0.1 & 12.6 $\pm$ 0.8 & ~ & 59.7 $\pm$ 0.8 & 49.6 $\pm$ 0.1 & 12.4 $\pm$ 0.3 \\ 
            \lwf & \bf 62.2 $\pm$ 1.8 & 49.1 $\pm$ 0.2 & 11.1 $\pm$ 1.2 & ~ & 55.8 $\pm$ 1.4 & 48.7 $\pm$ 0.1 & 12.5 $\pm$ 0.7 & ~ & 58.0 $\pm$ 1.2 & 49.2 $\pm$ 0.1 & 12.6 $\pm$ 0.6 \\ 
            \er & \bf 45.0 $\pm$ 24.7 & 23.8 $\pm$ 19.1 & 3.0 $\pm$ 2.5 & ~ & 70.0 $\pm$ 7.4 & 48.9 $\pm$ 0.3 & 7.6 $\pm$ 1.5 & ~ & 77.6 $\pm$ 1.8 & 49.3 $\pm$ 0.1 & 6.9 $\pm$ 0.5 \\ 
            \der & 41.4 $\pm$ 13.1 & 19.6 $\pm$ 17.4 & 7.9 $\pm$ 1.7 & ~ & 76.2 $\pm$ 1.9 & 49.3 $\pm$ 0.1 & 7.3 $\pm$ 0.4 & ~ & 77.5 $\pm$ 3.6 & 49.5 $\pm$ 0.1 & 6.8 $\pm$ 0.9 \\ 
            \derpp & 36.3 $\pm$ 21.7 & 23.6 $\pm$ 18.0 & 3.7 $\pm$ 3.5 & ~ & \bf 76.8 $\pm$ 7.3 & 49.0 $\pm$ 0.6 & 5.9 $\pm$ 0.5 & ~ & \bf 82.1 $\pm$ 7.1 & 49.4 $\pm$ 0.2 & 5.0 $\pm$ 1.6 \\ 
            \method & 38.8 $\pm$ 26.3 & 24.1 $\pm$ 18.1 & 4.9 $\pm$ 2.0 & ~ & 67.9 $\pm$ 2.0 & \bf 77.67 $\pm$ 1.9 & \bf 53.2 $\pm$ 3.9 & ~ & 70.7 $\pm$ 2.1 & \bf 80.2 $\pm$ 1.9 & \bf 57.1 $\pm$ 3.9 \\ 
            \bottomrule
    \end{tabular}

%% file: tables/CLEVR-ALL-RUNS.tex
      \setlength{\tabcolsep}{3pt}

\begin{tabular}{lccccccccccc}
        \toprule 
        & \multicolumn{3}{c}{\sc Supervision $0 \%$ } & & \multicolumn{3}{c}{\sc Supervision $1 \%$ } & & \multicolumn{3}{c}{\sc Supervision $10 \%$ } \\ 
        \cmidrule{2-4} \cmidrule{6-8} \cmidrule{10-12}
         & \sc Class-IL($\vY$) & \sc Class-IL($\vC$) & \sc {OOD  ($\vY$)} & ~ & \sc Class-IL($\vY$) & \sc Class-IL($\vC$) & \sc {OOD  ($\vY$)} & ~ & \sc Class-IL($\vY$) & \sc Class-IL($\vC$) & \sc {OOD  ($\vY$)} \\ 
        \naive & 44.9 $\pm$ 0.7 & 9.2 $\pm$ 1.2 & 25.0 $\pm$ 0.9 & ~ & 43.5 $\pm$ 2.6 & 19.9 $\pm$ 2.6 & 27.9 $\pm$ 11.4 & ~ & 39.9 $\pm$ 1.6 & 17.8 $\pm$ 0.1 & 16.9 $\pm$ 3.6 \\ 
        \restart & 39.1 $\pm$ 0.9 & 10.5 $\pm$ 1.4 & 11.1 $\pm$ 3.7 & ~ & 39.9 $\pm$ 1.0 & 17.8 $\pm$ 0.1 & 14.8 $\pm$ 4.1 & ~ & 39.2 $\pm$ 1.1 & 17.9 $\pm$ 0.1 & 13.7 $\pm$ 3.8 \\ 
        \ewc & 41.5 $\pm$ 7.1 & 8.3 $\pm$ 3.2 & 15.7 $\pm$ 8.4 & ~ & 38.5 $\pm$ 6.4 & 17.2 $\pm$ 1.4 & 23.5 $\pm$ 8.6 & ~ & 41.8 $\pm$ 1.8 & 17.8 $\pm$ 0.1 & 16.8 $\pm$ 4.3 \\ 
        \lwf & 47.1 $\pm$ 0.8 & 6.6 $\pm$ 2.2 & 27.2 $\pm$ 2.8 & ~ & 41.9 $\pm$ 1.8 & 19.1 $\pm$ 2.9 & 29.0 $\pm$ 15.2 & ~ & 39.5 $\pm$ 7.5 & 17.4 $\pm$ 3.9 & 44.8 $\pm$ 24.3 \\ 
        \er & 85.3 $\pm$ 0.3 & 10.2 $\pm$ 5.5 & 26.3 $\pm$ 3.5 & ~ & 72.8 $\pm$ 4.9 & 20.5 $\pm$ 3.9 & 28.8 $\pm$ 9.9 & ~ & 66.8 $\pm$ 8.4 & 18.3 $\pm$ 1.3 & 23.3 $\pm$ 4.0 \\ 
        \der & 77.7 $\pm$ 1.4 & 7.7 $\pm$ 2.8 & \bf 27.5 $\pm$ 2.7 & ~ & 73.3 $\pm$ 3.0 & 19.1 $\pm$ 1.9 & 24.9 $\pm$ 3.4 & ~ & 75.1 $\pm$ 3.9 & 19.2 $\pm$ 2.0 & 26.1 $\pm$ 5.0 \\ 
        \derpp & \bf 85.5 $\pm$ 0.2 & 11.1 $\pm$ 4.4 & 23.2 $\pm$ 1.8 & ~ &  83.0 $\pm$ 1.1 & 20.9 $\pm$ 3.9 & 32.4 $\pm$ 12.1 & ~ & 81.3 $\pm$ 2.3 & 20.1 $\pm$ 3.9 & 28.4 $\pm$ 9.3 \\ 
        \method & 70.7 $\pm$ 5.7 & 8.5 $\pm$ 4.5 & 25.6 $\pm$ 2.3 & ~ & \bf 83.2 $\pm$ 0.5 & \bf 85.9 $\pm$ 2.9 & \bf 91.1 $\pm$ 1.5 & ~ & \bf 85.2 $\pm$ 0.3 & \bf 87.9 $\pm$ 0.1 & \bf 91.9 $\pm$ 0.2 \\ \bottomrule
\end{tabular}